\documentclass[nojss]{jss}
\usepackage{hyperref}
\usepackage{multirow}
\usepackage{mathtools}
\DeclarePairedDelimiter{\ceil}{\lceil}{\rceil}
\usepackage{graphicx} 
\usepackage{subfigure}
\usepackage{amsmath,amssymb }
\usepackage[ruled] {algorithm2e}
\usepackage{array}
\usepackage{rotating}
\newcolumntype{L}[1]{>{\raggedright\let\newline\\\arraybackslash\hspace{0pt}}m{#1}}
\newcolumntype{C}[1]{>{\centering\let\newline\\\arraybackslash\hspace{0pt}}m{#1}}
\newcolumntype{R}[1]{>{\raggedleft\let\newline\\\arraybackslash\hspace{0pt}}m{#1}}

\newcolumntype{M}{>{\centering\arraybackslash}m{.44\linewidth}}

\newcommand{\KG}{\text{KG}}
\newcommand{\IE}{\text{IE}}

\newcommand{\UCB}{\text{UCB}}
\newcommand{\UCBE}{\text{UCB-E}}
\newcommand{\UCBV}{\text{UCB-V}}
\newcommand{\Kriging}{\text{Kriging}}

\author{Yingfei Wang\\Princeton University \And 
        Warren Powell\\Princeton University}
\title{ \pkg{MOLTE}: a Modular Optimal Learning Testing Environment}

\Plainauthor{Yingfei Wang, Warren Powell} 
\Plaintitle{MOLTE : a Modular Optimal Learning Testing Environment} 
\Shorttitle{\pkg{MOLTE}: a Modular Optimal Learning Testing Environment} 

\Abstract{
   We  address the relative paucity of empirical testing of learning algorithms (of any type) by introducing a new public-domain, Modular, Optimal Learning Testing Environment (\pkg{MOLTE}) for  Bayesian ranking and selection problem, stochastic bandits  or sequential experimental design problems.  The Matlab-based simulator allows the comparison of a number of learning
policies (represented as a series of .m modules) in the context of a wide range of
problems (each represented in its own .m module)  which makes it easy to add new algorithms and new test problems. State-of-the-art policies and various problem classes are provided in the package. The choice of problems and policies is guided through a spreadsheet-based interface.  Different graphical metrics are included. \pkg{MOLTE} is designed to be compatible with parallel computing to scale up from local desktop to clusters and clouds. We offer \pkg{MOLTE} as an easy-to-use tool for the research community that will make it possible to perform much more comprehensive testing, spanning a broader selection of algorithms and test problems.  We demonstrate the capabilities of \pkg{MOLTE} through a series of comparisons of policies on a starter library of test problems.  We also address the problem of tuning and constructing priors that have been largely overlooked in optimal learning literature. We envision \pkg{MOLTE} as a modest spur to provide researchers an easy environment  to study interesting questions involved in optimal learning.
}
\Plainkeywords{Bayesian optimization, ranking and selection, multi-armed bandits, sequential decision making, Matlab} 

\Address{
  Yingfei Wang\\
  Department of Computer Science\\
  Princeton University\\
  Princeton, NJ, 08540\\
  E-mail: \email{yingfei@cs.princeton.edu}\\
  ~\\
  Warren Powell\\
  Department of Operations Research and Financial Engineering\\
  Princeton University\\
  Princeton, NJ, 08544\\
  E-mail: \email{powell@princeton.edu}  
}

\begin{document}

\section{Introduction}
We consider sequential decision problems in which at each time step, we choose one of finitely many alternatives and observe a random reward. The rewards are independent of each other and follow some unknown probability distribution. One goal can be to identify the alternative with the best expected performance within a limited measurement budget, which is the objective of offline ranking and selection. Another goal can be to maximize the expected cumulative sum of rewards obtained in a sequence of allocations, a problem class often addressed under the umbrella of  multi-armed bandit problems. 

Ranking and selection problems and/or multi-armed bandits arise in many settings. We
may have to choose a type of material that has the best performance,
the features in a laptop or car that produce the
highest sales, or the molecular combination that produces
the most effective drug. In health services, 
physicians have to make medical decisions (e.g. a course of drugs, surgery, and expensive tests) to provide the best treatment. In online advertisements, the system would like to choose ads to gather the most ad-clicks. 

Since the seminal paper by \cite{lai1985asymptotically}, there has been a long history in the optimal learning literature of proving some sort of bound, supported at times by relatively thin empirical work by comparing a few policies on a small number of randomly generated problems \citep{audibert2010best,cappe2013kullback,srinivas2009gaussian,auer2002finite,garivier2008upper,wang2016finite, Audibert:2009:ETU:1519541.1519712}.  The problem, of course, is that compiling a library of test problems, and then running an extensive set of comparisons, is difficult.  The problem is this means that we are analyzing the finite time performance of algorithms using bounds that only apply asymptotically by limited empirical experiments to support the claim of finite time performance. To this end,  we introduce a new Modular Optimal Learning Testing Environment (\pkg{MOLTE}) for comparing a number of policies on a wide range of learning problems, providing the most comprehensive testbed that has yet appeared in the literature.
 
Similar libraries have been proposed for  Bayesian optimization in different programming languages with  different metrics and visualizations, for example, \pkg{BayesOpt} \citep{martinez2014bayesopt} and  \pkg{Spearmint} \citep{snoek2012practical}.   Yet the uniqueness of \pkg{MOLTE} lies in its design goal to  facilitate  comprehensive comparisons, on a broader set of test problems and a broader set of policies (which is not to restricted to Bayesian algorithms), rather than just a code library.  With its unique  modular design, \pkg{MOLTE} allows users to easily specify their own problems or their own  algorithms without limitation as long as they follow the general function interface.  The choice of problems and policies is guided through a spreadsheet-based interface. Since many of the algorithms have tunable parameters, we include the feature that the user can easily indicate in the spreadsheet to specify the value of the tunable parameter, or ask the package to optimize the tunable parameter. We have designed various (graphical) comparison metrics in order to gain a comprehensive understanding of different policies from different perspective.   \pkg{MOLTE} is also designed to be compatible with parallel computing to scale up from local desktop to clusters and clouds.  We offer \pkg{MOLTE} as an easy-to-use tool for the research community  that provides a highly flexible environment for testing a range of learning policies on a library of test problems,  so that researchers can more easily draw insights into the behavior of different policies in the context of different problem classes. 

\pkg{MOLTE} is designed for problems where decisions can be represented as a set of discrete alternatives.  These might be materials, drug combinations, features in a product, and medical decisions.  They might also be discretized continuous decisions such as temperatures, pressures, concentrations, length and time (e.g. how long a material is soaked in a bath). If there are more than two or three dimensions, it is possible to use a sampled set of alternatives (this set can be updated from time to time, but \pkg{MOLTE} is designed to handle a single set of alternatives).

This paper is organized as follows. In Section \ref{model}, we lay out the mathematical models for sequential decision problems. Section \ref{Imp} describes how \pkg{MOLTE} is implemented and describes the package from a user's perspective.   We demonstrate the ability of \pkg{MOLTE} and the types of reports that it produces through extensive experimental results in Section \ref{rs} and \ref{bandit}. We draw the conclusion that there is no universal policy that works the best under all problem classes and the existence of bounds does not appear to provide reliable guidance regarding which policy works best.   In practice, we believe that more useful guidance could be obtained by abstracting a real world problem, running simulations and using these to indicate which policy works best. We envision \pkg{MOLTE} as a modest spur to induce other researchers to come forward to study interesting questions involved in optimal learning, for example, the issue of tuning  as discussed in Section \ref{Dis}.

\section{Sequential decision problems}\label{model}
Suppose we have a collection $\mathcal{X}$ of $M$ alternatives, each of which can be measured sequentially to estimate its unknown performance $\mu_x=\mathbb{E}[F(x,W)]$. The utility function $F(x,W)$ can be understood as costs, rewards or losses, where $x \in \mathcal{X}$ is a decision variable and $W$ is a random variable.  The initial state $S^0$ is used to capture all information given as prior input.  At each time step $n$, we use some policy to choose one alternative to measure $x^n = X^{\pi}(S^n)$ and receive a stochastic reward $\hat{F}^{n+1}=F(x^n, W^{n+1})$. After the decision and information, the system transitions to the state of belief at the next point in time according to some known transition function $S^{n+1} = S^{M}(S^n, x^n, \hat{F}^{n+1})$.  Two styles of objective functions are considered in this paper:
\begin{itemize}
\item{Terminal reward} -- considered in Bayesian optimization, ranking and selection problems, and also known as simple regret in multi-armed bandits.  Here we assume have a limited budget of N function evaluations which have to be sequentially allocated over the different alternatives $x\in\mathcal{X}$ using a policy $\pi$.  We use this policy to produce estimates $\theta^{\pi,N}_x$ of $\mu_x$, and then choose the best design:
\begin{equation*}
X^{\pi,N} = \arg\max_x \theta^{\pi,N}_x.
\end{equation*}
We can state the problem of finding the best experimental policy as
\begin{equation}\label{offline}
\max_{\pi}\mathbb{E}\Big[F(X^{\pi,N},W)|S^0\Big].
\end{equation}
 In this case, we are not punished for errors incurred during training and instead are only concerned
with the final recommendation after the offline training phases. It should be noted that the expectation is over  different sets of random variables.  
The first is the sequence of observations $W^1, \dots, W^{N}$ which then produces the random $X^{\pi,N}$. The second expectation is over $W$ in the equation, which is used to evaluate the solution. If a Bayesian approach is used, there is a third level of expectation over the prior.  
\item{Cumulative reward} -- extensively studied under the umbrella of multi-armed bandits. If we have to experience the rewards while we do our learning/exploring, we may want to maximize contributions over some time horizon. The (online) objective function would be written as 
\begin{equation}\label{online}
\max_{\pi}\mathbb{E}\Big[\sum_{n=0}^{N-1}F(X^{\pi}(S^n),W^{n+1})|S^0\Big],
\end{equation}
where the expectation is over the sequence of observations $W^1, \dots, W^{N}$ and the prior if any.
\end{itemize}

Despite different styles of  objective functions, a general algorithm for sequential decision problems can be summarized in Algorithm \ref{al}:
\begin{algorithm}\label{al}
\caption{General algorithm for sequential decision problems}
\SetKwInOut{Input}{input}\SetKwInOut{Output}{output}
 \Input{time horizon $N$, initial state $S^0$, policy $\pi$, transition function $S^M$} 
 \For{$n=0$ to $N$}{
Select the point $x^n = X^{\pi}(S^n)$\\
Observe $\hat{F}^{n+1}=F(x^n, W^{n+1})$\\
Update the state $S^n = S^{M}(S^n, x^n, \hat{F}^{n+1})$}
\end{algorithm}

It should be noted that the states $S^n$ can be understood as belief states which specifies the posterior on the unknown function $F(x,\cdot)$ if a Bayesian approach is used as in Bayesian optimization and ranking and selection problems. The transition function $S^M$ then follows from Bayes' Theorem. We assume a normally
distributed prior with heteroscedastic measurement noise (that is, the variance of an experiment depends on $x$),  with known  standard deviation $\sigma_x^W$.  We begin with a normally distributed Bayesian prior belief $\mu_x \sim \mathcal{N}(\theta^0, \Sigma^0)$. For convenience, we introduce the $\sigma$-algebras  $\mathcal{F}^{n}$ for any $n =0,1,...,N-1$ which is formed by the previous $n$ measurement choices and outcomes, $x^0, W^1,..., x^{n-1},W^n$.  We define $\theta^n_x=\mathbb{E}[\mu_x|\mathcal{F}^n]$ and $\Sigma^n$ the conditional covariance matrix.  After a measurement $W^{n+1}$ of alternative $x$, a posterior
distribution on the beliefs are calculated by:
\begin{align}\label{aabb}
    \theta^{n+1} &= \Sigma^{n+1}\left( \left(\Sigma^n\right)^{-1}\theta^n + \beta^W W^{n+1} e_x\right), \\ \label{bb}
    \Sigma^{n+1} &= \left( \left(\Sigma^n\right)^{-1} + \beta^W e_x e_x^T\right)^{-1},
\end{align}
where $e_x$ is the vector with 1 in the entry corresponding to alternative $x$ and 0 elsewhere. $S^n=(\theta^n, \Sigma^n)$ is then our state of knowledge in this case.

There are other models that go beyond the normal-normal model studied in the work. Examples include logistic models \citep{chapelle2011empirical,wang2016knowledge}, contextual bandits \citep{agrawal2012thompson, chu2011contextual, zhangepoch} and bandit with multi-plays \citep{wang2017efficient, slivkins2013ranked}.

\section{Software implementation}\label{Imp}
\label{sec_numerical}

In this section, we describe the implementation of \pkg{MOLTE}\footnote{The software is available at \url{http://www.castlelab.princeton.edu/software.htm}.} that is designed to test a variety of different learning policies on a library of test problems.  The architecture makes it particularly easy for researchers to add new policies, and new problems.
\subsection{Structural overview}
\pkg{MOLTE} is a Matlab-based modular architecture, where policies and problems are captured in a set of .m files, which makes it easy for researchers to add new policies and new problems. \code{MOLTE.m} compares the polices specified in an Excel spreadsheet for each problem
class for \code{numP} times. Each time the
simulator is run, it generates \code{numTruth}  different
sample paths, shared between all the policies, computes the value of the objective
function for each sample path and then averages the \code{numTruth} replicas as the expected {\it terminal reward}
or the expected {\it cumulative rewards}. The user may select in the spreadsheet to evaluate policies
using either an online (cumulative reward) objective function Eq. \eqref{online}, or an offline (terminal reward) objective function Eq. \eqref{offline}
(ranking and selection, Bayesian optimization).

In order to speed up the comparison,  \pkg{MOLTE} is specially designed to be compatible with parallel computing to scale up from local desktop to clusters and clouds. This can be achieved by first invoking \code{matlabpool} to submit a batch job to start a parallel environment and then use \code{parfor i=1:numP} instead of \code{for i=1:numP} on Line 111. 

 We pre-coded  a number of standard test functions, including
  \begin{itemize}
  \item problems with additive Gaussian noise, such as
  	\begin{itemize}
	 \item  Branin's function  \citep{dixon1978global}, 
	 \item Goldstein Price function  \citep{hu2008model}, 
	 \item Rosenbrock function  \citep{hu2008model}, 
	 \item Griewank function \citep{hu2008model},
	 \item six-hump camel back function by \citep{molga2005test}),
	 \end{itemize} 
  \item synthetic bandit experiments \citep{audibert2010best}, 
  \item Gaussian process regression,
  \item real-world applications like newsvendor problems and payload delivery. 
  \end{itemize}We also pre-coded a number of competing policies, such as 
  \begin{itemize}
  \item UCB variants \citep{auer2002finite}, 
  \item successive rejects, 
  \item sequential Kriging (SKO, as a representative of Bayesian global optimization \citep{jones1998efficient, huang2006global, jones2001taxonomy}), 
  \item Thompson sampling,
  \item the knowledge gradient policies \citep{frazier2008knowledge}. 
  \end{itemize} Each of the problem classes and policies is organized in its own Matlab file, so that  it is easy for a user to add in a new problem or a policy.  In order to make a fair comparison, all the observations are pre-generated and shared between competing policies. There may be problems where a domain expert can provide prior knowledge (such as the likely success of a drug for a particular patient, the popularity of an online movie, or the sales potential of a set of features in a laptop), but in some settings these have to be derived from data.  In \pkg{MOLTE} we provide various ways to construct a prior, including 
  \begin{itemize}
  \item user-provided prior distributions, 
  \item hard-coded default prior distributions, 
  \item an uninformative prior,
  \item maximum likelihood estimation MLE (see Section \ref{PG}). 
  \end{itemize}

\subsection{Input Arguments}
The input to the simulator is an Excel spreadsheet \code{ProblemsandAlgorithms.xls} which  allows users to specify  the problem classes and competing policies, as well as the belief models, the objectives, the prior construction and the measurement budgets.   We provide a sample input spreadsheet in Table \ref{input}.  For policies that have tunable parameters, a star included in the parentheses after the policy will initiate an automatic brute force tuning procedure with the optimal value reported in \code{alpha.txt}.  The logic anticipates that tunable parameters may be anywhere from $10^{-5}$ up to $10^{5}$. Whereas the user can also specify the value to be used for the policy in the parentheses. 
\begin{table}[htp!]
\small
\setlength{\tabcolsep}{0.01pt}
\centering
\caption{Sample input spreadsheet.}\label{input}
\begin{tabular}{|C{2cm}|c|C{2.4cm}|c|C{1.4cm}|C{1.8cm}|c|c|c|c|}
    \hline
\textbf{Problem class}&\textbf{Prior} & \textbf{Measurement Budget}&\textbf{Belief Model}	& \textbf{Offline/ Online}  &\textbf{Number of policies} & 		 & 		 & 		    & \\ \hline
Bubeck1   		       	 & Uninform   & 10    					& independent	& Online  		& 3   				& OLKG & IE(*)	 & UCB      &  \\\hline
Branin     	    			 & MLE     		  & 5 					& independent	&  Offline 		&  4  				&UCBE(*)&IE(1.7)&KG &SR   \\ \hline
GPR      					 &Default            &0.3                   & correlated     &  Online         &  4                  &KLUCB &EXPL  &UCB&TS  \\ \hline
NanoDesign			 &MLE			&0.5					 &correlated      & Offline         &3                     &Kriging &EXPT  &KG  &\\ \hline
\end{tabular}
\label{rosen}
\end{table}

\textbf{Problem class} is the name of a pre-coded problem with a specified truth function, the
number of alternatives and a default noise level. If it is a user defined problem, the user
should write a .m file in the \code{./problemClasses} folder with the same name as presented in
this spreadsheet. Due to the high popularity of Gaussian Process Regression (GPR), we offer the flexibility of  directly specify the  values of the parameter of  GPR in the spreadsheet. For example, GPR($\sigma,
\beta;M$) specifies the value of the parameters as follows \citep{powell2012optimal}: the prior mean $\theta^0_x$ is drawn from $\mathcal{N}(0, \sqrt\sigma)$, the prior covariance matrix $\Sigma^0$ is of
the form $\sigma\exp(-\beta(x-x'))$ and $M$ is the number of alternatives.

\textbf{Prior} indicates the ways to get a prior. 
\begin{itemize}
\item{\it MLE} means using Latin hypercube designs and
MLE for initial fit. \item{\it Default} can be used only for the problems (e.g. GPR and
InanoparticleDesign) that have a default prior. \item{\it Given} means using the prior distribution
provided by the user. It can be achieve either by specifying the parameters of the
problem class, e.g. GPR(50, 0.45;100), or by providing a \code{Prior_problemClass.mat} file
containing \code{mu_0}, \code{covM} and \code{beta_W} in the \code{./Prior} folder, e.g. \code{Prior_GPR.mat}.
\item{\it Uninformative} specifies  mean zero and infinite variance for each alternative.
\end{itemize}

\textbf{Measurement Budget} specifies the ratio between the time horizon of the decision
making procedure to the number of alternatives. For example, in the spreadsheet a 5 means that the horizon will be 5 times the number of alternatives (which is 100), producing a total experimental budget of 500.

\textbf{Belief Model}  specifies whether we are using independent or correlated beliefs for the
policies which use a Bayesian belief model.

\textbf{Offline/Online} controls whether the objective is to maximize the expected terminal  reward Eq. \eqref{offline}
or the expected cumulative rewards Eq. \eqref{online}.

\textbf{Number of Policies} is the number of policies under comparison. This specifies the
number of columns which contain the name of a policy to be tested, each represented in the corresponding \code{.m} file with the same name. If there are parentheses with a number
after the name of the policy, it means setting the tunable parameter to the value specified
in the parentheses. If there are parentheses with $*$, it means tuning the parameters and using the tuned value in the comparison; otherwise use the
default value (in fact some policies, e.g. KG and Kriging, do not have tunable
parameters).  All policies are compared against the first policy in the list.

\subsection{Output}
All the data and figures are saved in a separate folder for each problem class. Within the
folder of each problem class, each one of the \code{numP} folders (with the folder name from 1 to \code{numP}) contains:

\code{objectiveFunction.mat} saves the value of the online or offline objective function 
achieved by each policy for each of the \code{numP} replica.

\code{choice.mat} saves the decisions made by each policy in a variable named \code{choices} and the name of all policies in   another variable \code{policies}. This file is only obtained for the first trial so that the dimension of \code{choices} is  $number\_of\_policies \times M \times$ \code{numTruth}.

\code{FinalFit.mat} saves the final estimate of the surface by each policy after the
measurement budget exhausted, together with the corresponding truth. This file is only obtained for the first trial.

 \code{alpha.txt} saves the value of tunable parameter for each policy that requires tuning, i.e. with a ($*$) in the input spreadsheet.
 
\code{offline_hist.pdf} is the histogram for each policy describing the distribution over \code{numP} trials of the
expected terminal reward compared to the reward obtained by the reference policy
(which is the first policy in the input spreadsheet).

\begin{figure}[htp!]
    \centering
        \includegraphics[width=0.5\textwidth]{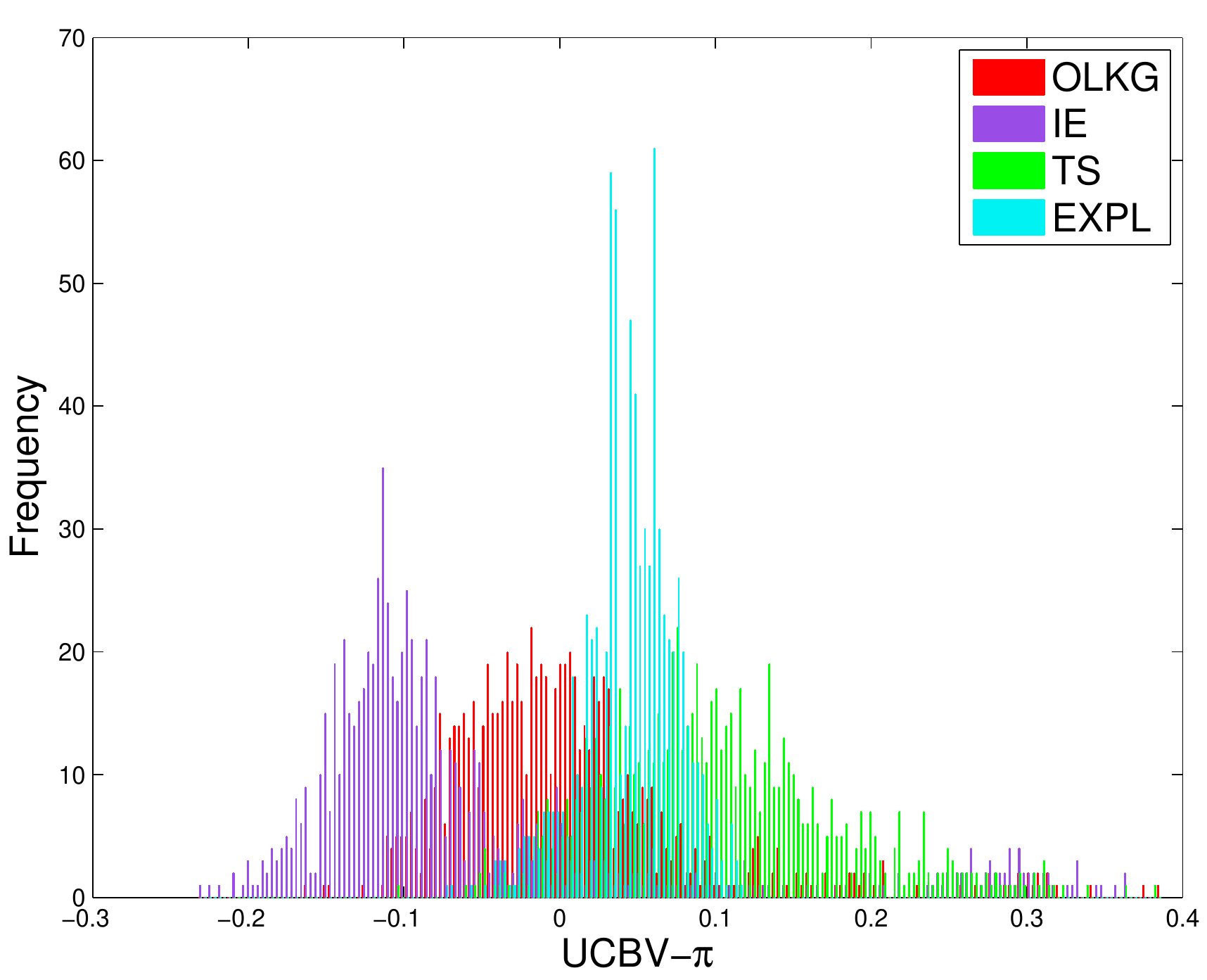}
 	\caption{Example figure of online\_hist.pdf. \label{ex}}
 \end{figure}

\code{online_hist.pdf} is the histogram describing the distribution  of the expected cumulative
reward over \code{numP} trials. One of the example figure is Fig. \ref{ex}. A distribution
centered around a positive value implies the policy underperforms the reference policy, which in this example is UCBV.
 
\begin{figure}[htb]
    \centering
        \includegraphics[width=0.5\textwidth]{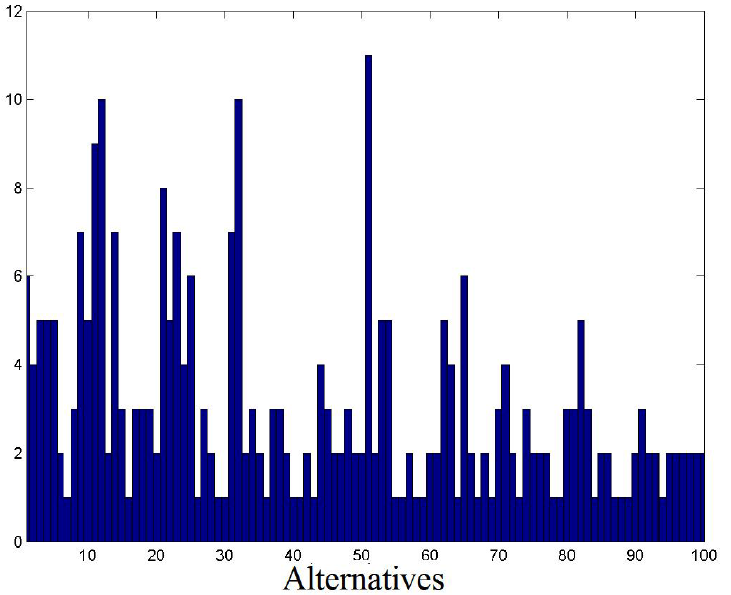}
 	\caption{Example figure of the histogram of the frequency of choosing each of the alternative under a policy. \label{exF}}
 \end{figure}

It is always useful for researchers to examine the sampling pattern of each policy to gain a better understanding of its behavior. To this end, we provide a function \code{histChoice.m} that reads in the \code{choice.mat} and generates the distribution of the frequency of choosing each of the alternatives for each policy. \code{filedir} specifies which one of the \code{numP} trials is used to generate the sampling pattern, e.g. \code{filedir='./1/'}. Since within each trial, \code{numTruth} different truths are sampled,  \code{numT} is used to indicate the number of truths the user would like to draw the sampling pattern from. Fig. \ref{exF}  is an example of a sampling pattern with the x-axis the 100 alternatives and the histogram of the sampling pattern under a measurement budget of 300.

We also provide other graphical metrics for comparing the policies. \code{genProb.m} can read in the \code{objectiveFunction.mat} and depict the mean opportunity cost
with error bars indicating the standard deviation of each policy as shown
in Fig. \ref{xl}, together with the probability of each policy being optimal
and being the best in Fig. \ref{xr}:

\begin{figure}[htp!]
    \centering
      \subfigure[Opportunity cost]{   \includegraphics[width=0.44\textwidth]{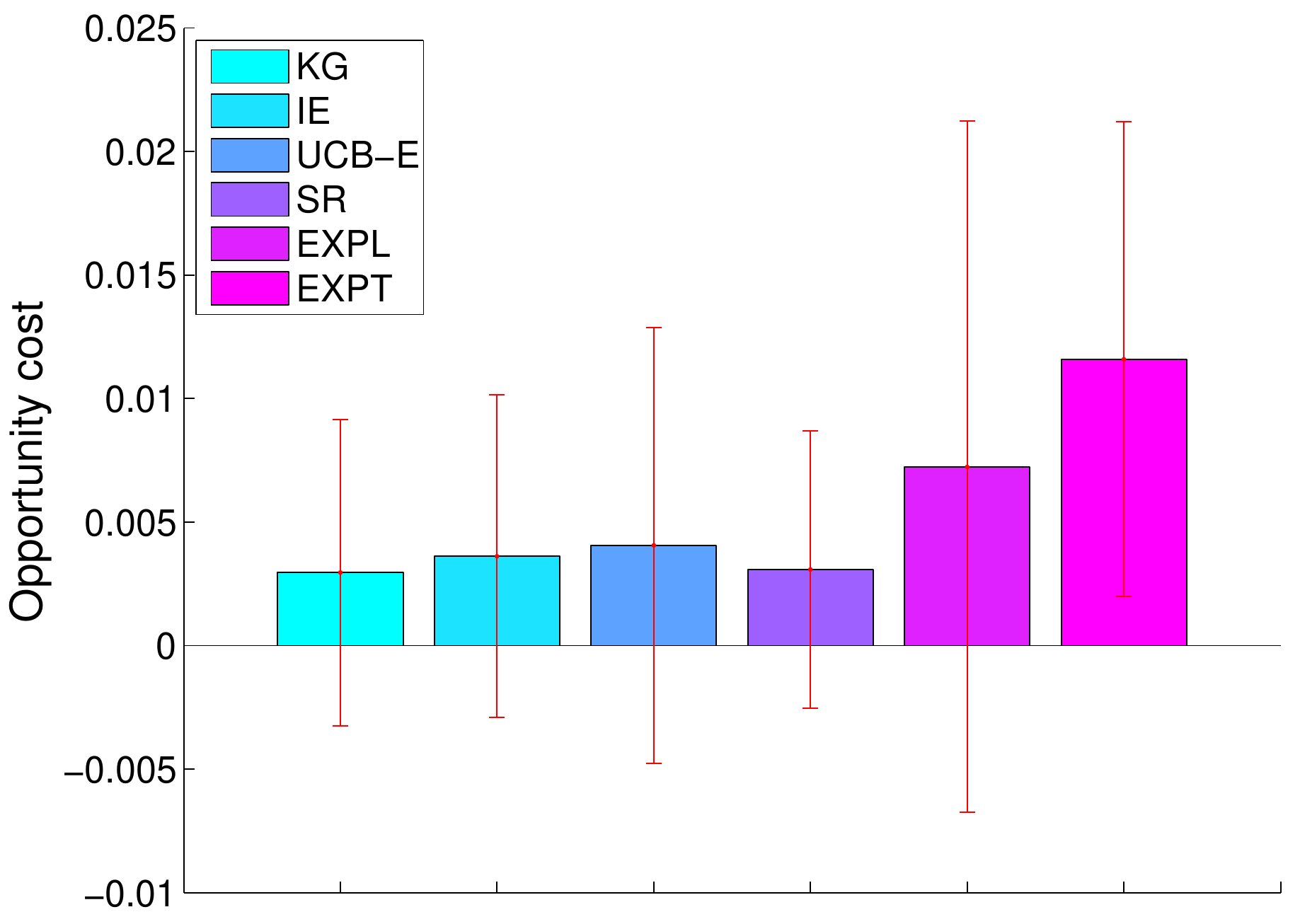} \label{xl}}~~~
   \subfigure[Probability of optimality/winning]{\includegraphics[width=0.445\textwidth]{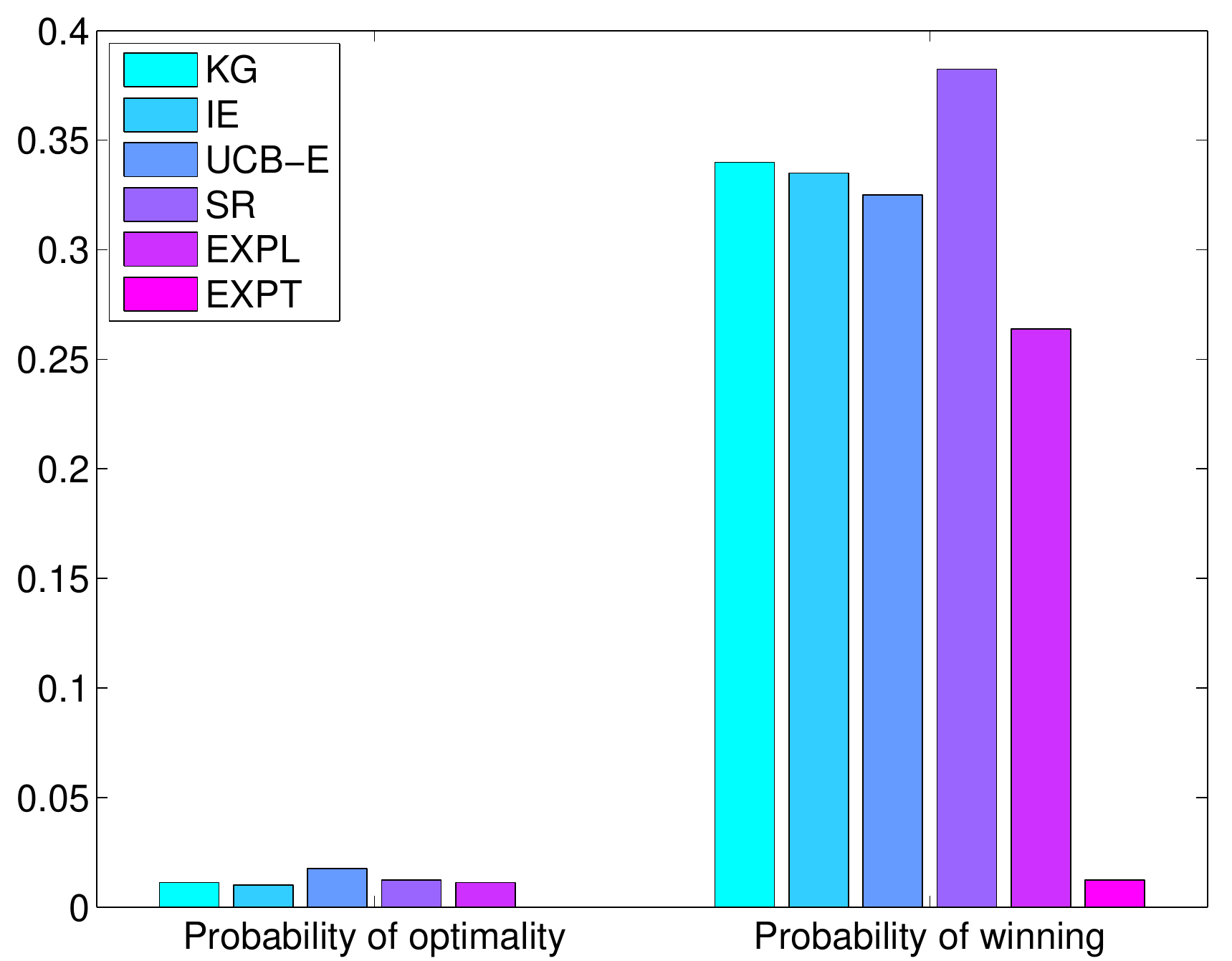}\label{xr}}
    \caption{ (a)  depicts the mean opportunity cost with error bars indicating the standard deviation of each policy. The first bar group in (b) demonstrates the probability that the final recommendation of each policy is the optimal one.  The second bar group in (b)  illustrates the probability that the opportunity cost of each policy is the lowest. }
\end{figure}

The statistics stored in \code{objectiveFunction.mat}, \code{choice.mat} and \code{FinalFit.mat} can easily
be used for other illustrations. For example, one can use the truth values stored in
\code{FinalFit.mat} and the number of times each policy samples each alternative in 
\code{choice.mat} to generate two dimensional contour plot using Matlab
commands \code{contour(\dots)}, \code{plot(\dots)} and \code{text(\dots)}, as well as the corresponding posterior
contour using the final estimate of the surface stored in \code{FinalFit.mat}, as we demonstrate later in Fig. \ref{lr}.

\subsection{Pre-coded problem classes} \label{pc}

While a wide range of problem classes and policies are precoded in \pkg{MOLTE}, in the next two subsections we only briefly summarize the problem classes and policies  mentioned in the following numerical experiments of this paper.  As of this writing,  \pkg{MOLTE} includes 23 pre-coded problem classes, and 20 pre-coded policies.

\textbf{Bubeck's Experiments:} \citep{audibert2010best}
We consider Bernoulli distributions with the mean of the best arm always $\mu_1=0.5$. $M$ is the number of arms.\\
\textbf{Bubeck1:} $M=20$, $\mu_{2:20}=0.4$.\\
\textbf{Bubeck2:} $M=20$, $\mu_{2:6}=0.42$, $\mu_{7:20}=0.38$.\\
\textbf{Bubeck3:} $M=4$, $\mu_i=0.5-(0.37)^i$, $i \in \{2,3,4\}$.\\
\textbf{Bubeck4:} $M=6$, $\mu_2=0.42$, $\mu_{3:4}=0.4$, $\mu_{5:6}=0.35$.\\
\textbf{Bubeck5:} $M=15$, $\mu_i=0.5-0.025i, i\in\{2,\cdots, 15\}$. \\
\textbf{Bubeck6:} $M=20$, $\mu_2=0.48$, $\mu_{3:20}=0.37$. \\
\textbf{Bubeck7:} $M=30$, $\mu_{2:6}=0.45$, $\mu_{7:20}=0.43$, $\mu_{21:30}=0.38$.

\textbf{Asymmetric unimodular function (AUF):} $x$ is a controllable parameter ranging from 21 to 120.  The objective function is $F(x,\xi)=\theta_1\min(x,\xi)-\theta_2x$,  where $\theta_1$,  $\theta_2$ and the distribution of the random variable $\xi$ are all unknown. $\xi$ is taken as a normal distribution with mean 60.  Three noise levels  are considered by setting different noise ratios between the standard deviation and the mean of $\xi$: HNoise--0.5,  MNoise--0.4, LNoise--0.3. Unless explicitly pointed out, experiments are taken under LNoise.

\textbf{Equal-prior: } $M=100$. The true values $\mu_x$ are uniformly distributed over $[0, 60]$ and measurement noise $\sigma_W=100$. $\theta_x^0=30$ and $\sigma_x^0=10$ for every x.

All the standard optimization test functions  are flipped in MOLTE to generate maximization problems instead of minimization in line with  R\&S and bandit problems. The standard deviation of the additive Guassion noise is set to $20$ percent of the range of the function values. 

\textbf{Rosenbrock functions with additive noise:}
$$
f(x,y, \phi) = 100(y-x^2)^2+(1-x)^2+\phi,
$$
where $-3\le x\le 3$, $-3\le y \le 3$.  x and y are uniformly discretized into 13 $\times$ 13 alternatives.

\textbf{Pinter's function with additive noise:} 
\begin{align} \nonumber
f(x, y,\phi)&= \log_{10}\big{(}1+(y^2-2x+3y-\cos x+1)^2\big{)}+ \log_{10}\big{(}1+2(x^2-2y+3x-\cos y+1)^2\big{)} \\ \nonumber
            &+
            x^2+2y^2+ 20\sin^2(y\sin x-x+\sin y)+40\sin^2(x\sin y-y+\sin x) + 1+\phi,
\end{align}
where $-3\le x\le 3$, $-3\le y \le 3$. x and y are uniformly discretized into 13 $\times$ 13 alternatives.

\textbf{Goldstein-Price's function with additive noise:}  \begin{eqnarray*}
f(x,y,\phi) &=& [1 + (x + y + 1)^2(19 - 14x + 3x^2-14y+6xy+3y^2)]\cdot\\
&&[30 + (2x - 3y)^2
(18-32x + 12x^2+48y-36xy+27y^2)]+\phi,
\end{eqnarray*}
where $-3\le x\le 3$, $-3\le y \le 3$. x and y are uniformly discretized into 13 $\times$ 13 alternatives.

\textbf{Branins's function with additive noise:}
$$
f(x,y,\phi)=(y-\frac{5.1}{4\pi^2}x^2+\frac{5}{\pi}x-6)^2+10(1-\frac{1}{8\pi})\cos(x)+10+\phi,
$$
where  $-5\le x\le 10$, $0\le y \le 15$. x and y are uniformly discretized into 15 $\times$ 15 alternatives.

\textbf{Ackley's function with additive noise:} 
$$
f(x,y,\phi)=-20\exp\Big{(}-0.2\cdot\sqrt{\frac{1}{2}(x^2+y^2)}\Big{)}-\exp\Big{(}\frac{1}{2}(\cos(2\pi x)+\cos(2\pi y))\Big{)}+20+\exp(1)+\phi,
$$
where $-3\le x\le 3$, $-3\le y \le 3$. x and y are uniformly discretized into 13 $\times$ 13 alternatives.

\textbf{ Hyper Ellipsoid function with additive noise:}$$
f(x,y,\phi)=x^2+2y^2+\phi.
$$
where $-3\le x\le 3$, $-3\le y \le 3$. x and y are uniformly discretized into 13 $\times$ 13 alternatives.

\textbf{ Rastrigin function with additive noise:}
$$
f(x,y,\phi)=20+\big{[}x^2-10\cos(2\pi x)\big{]}+\big{[}x^2-10\cos(2\pi y)\big{]}+\phi,
$$
where  $-3\le x\le 3$, $-3\le y \le 3$.  x and y are uniformly discretized into 11 $\times$ 11 alternatives.

 \textbf{ Six-hump camel back function with additive noise:} 
$$
f(x,y,\phi)=(4-2.1x^2+\frac{x^4}{3})x^2+xy+(-4+4y^2)y^2+\phi,
$$
where $-2\le x\le 2$, $-1\le y \le 1$.  x and y are uniformly discretized into 13 $\times$ 13 alternatives.

\subsection{Pre-coded policies}\label{tcp}
We have pre-coded various state-of-the-art policies $\pi$, which differ according to their decision
$X^{\pi, n}(S^n)$ of the alternative to measure at time $n$ given state $S^n$.

\textbf{Knowledge gradient (KG):}  \citep{frazier2008knowledge,frazier2009knowledge}
This policy is designed for offline objective \eqref{offline}. Define the knowledge gradient as 
\begin{equation*}\label{TKG}
\nu_x^{\KG,n}=\mathbb{E}[\max_{x'}\theta_{x'}^{n+1}-  \max_{x'} \theta^{n}_{x'}| x^n=x, S^n].
\end{equation*}
$$X^{\KG,n}(S^n)=\arg \max_{x \in \mathcal{X}} \nu_x^{\KG,n}.$$

\textbf{Online knowledge gradient (OLKG):}   \citep{ryzhov2012knowledge}
$$X^{\text{OLKG},n}(S^n)=\arg \max_{x \in \mathcal{X}} \theta_x^n + (N-n)\nu_x^{\KG,n}.$$

\textbf{Interval Estimation (IE): } \citep{kaelbling1}
\begin{equation*} \label{IE}
X^{\IE,n}(S^n)=\arg \max_x  \theta^n_x + z_{\alpha/2} \sigma_x^n,
\end{equation*}
where $z_{\alpha/2}$ is a tunable parameter.

\textbf{Kriging:} \cite{huang2006global}

Let $x^*=\arg\max_x(\theta^n_x+\sigma^n_x)$, and then
\begin{equation*}
X^{\Kriging,n}(S^n)=\arg \max_x(\theta_x^n-\theta_{x^*}^n)\Phi(\frac{\theta_x^n-\theta_{x^*}^n}{\sigma_x^n})+\sigma_x^n\phi(\frac{\theta_x^n-\theta_{x^*}^n}{\sigma_x^n}),
\end{equation*}
where $\phi$ and $\Phi$ are the standard normal density and cumulative distribution functions.

\textbf{Thompson sampling (TS):} \citep{thompson1933likelihood}   $$X^{\text{TS},n}(S^n) = \arg\max_{x} \tilde{\theta}_x^n,$$ where $\tilde{\theta}_x^n \sim \mathcal{N}(\theta^n_x, \sigma^n_x)$  for independent beliefs or $\tilde{\theta}_x^n \sim \mathcal{N}(\theta^n, \Sigma^n)$ for correlated beliefs.

\textbf{UCB: } \citep{auer2002finite}
\begin{equation*} \label{UCBN}
X^{\UCB,n}(S^n) = \arg \max_x \hat{\theta}^n_x + \sqrt{\frac{2 V_x^n\log{n}}{N_x^n}},
\end{equation*}
where $\hat{\theta}_x^n$, $V_x^n$, $N_x^n$ are the sample mean of $\mu_x$, sample variance of $\mu_x$, and number of times $x$ has been sampled up to
time $n$, respectively.  The quantity $\hat{\theta}_x^0$ is initialized by measuring each alternative once. These are similarly defined in the following variants of UCB.

\textbf{UCB-E: } \citep{audibert2010best}
\begin{equation*} \label{UCB-E}
X^{\UCBE,n}(S^n) =\arg \max_x  \hat{\theta}^n_x + \sqrt{\frac{\alpha}{N_x^n}},
\end{equation*}
where $\alpha$ is a tunable parameter.

\textbf{UCB-V: } \citep{Audibert:2009:ETU:1519541.1519712}
\begin{align*}
X^{\UCBV,n}(S^n) &=\arg \max_x  \hat{\theta}^n_x  \nonumber + \sqrt{\frac{V_x^n \log n}{N_x^n}}+ 1.5 \frac{\log n}{N_x^n}.  \label{UCB}
\end{align*}

\textbf{SR:} \citep{audibert2010best} Let $A_1=\mathcal{X}$, $\overline{\log}(M)=\frac{1}{2}+\sum_{i=2}^M\frac{1}{i}$, $$n_m=\ceil[\Big]{\frac{1}{\overline{\log}(M)}\frac{n-M}{M+1-m}}.$$

For each phase $m=1,...,M-1$:
\begin{enumerate}
\item For each $x \in A_{m}$, select alternative $x$ for $n_m-n_{m-1}$ rounds.
\item Let $A_{m+1}=A_{m}\setminus \arg\min_{x \in A_m} \hat{\theta}_x.$
\end{enumerate}

\textbf{KLUCB:} \citep{cappe2013kullback} 
$$
X^{\text{KLUCB},n}(S^n) =\arg \max_x  \hat{\theta}^n_x + \sqrt{\frac{2V_x^n (\log n+3\log \log(n))}{N_x^n}}. 
$$

\textbf{EXPL:} A pure exploration strategy that tests each alternative  equally often through random sampling of the set of alternatives.

\textbf{EXPT:} A pure exploitation strategy. $$X^{\text{EXPT},n}(S^n) = \arg \max_{x} \hat{\theta}^n_x.$$

\subsection{Prior Generation}\label{PG}
 MOLTE features the following strategies for building a prior:
\begin{itemize}
\item If an {\it uninformative} prior is specified by the user for independent beliefs, a uniform prior will be used with $\theta_x^0=0$ and $\sigma^0_x=
inf$ for every $x$. In such case, same as with frequentist approaches (for example, UCBs), Bayesian approaches  will measure each alternative once at the very beginning.  
\item User-defined priors can be   achieved either by specifying the parameters of the
problem class, e.g. GPR(50, 0.45;100), or by providing a \code{Prior_problemClass.mat} file
containing \code{mu_0}, \code{covM} and \code{beta_W} in the \code{./Prior} folder, e.g. \code{Prior_GPR.mat}.

\item If  maximum likelihood estimation ({\it MLE}) is chosen to obtain the prior distribution for either independent beliefs or correlated beliefs, we follow \cite{jones1998efficient} and \cite{huang2006global} to use Latin hypercube designs for initial fit. For independent beliefs, we adopt a uniform prior with the same mean value $\theta_x^0$ and standard deviation $\sigma_x^0$ for all alternatives. For  correlated beliefs, we use a constant mean value $\theta_x^0$ for all alternatives and  a prior covariance matrix of the form $$\Sigma_{xx'}^0=\sigma e^{-\sum_{i=1}^{d}\lambda_i(x_i-x_i')^2},$$
where each arm $x$ is a $d$-dimensional vector and $\sigma, \lambda_i$ are constant.   We adopt the rule of thumb by \cite{jones1998efficient} for the default number ($10\times p$) of points, where $p$ is the number of parameters to be estimated. In addition, as suggested by \cite{huang2006global}, to estimate the random errors, after the first $10 \times p$ points are evaluated, we add one replicate at each of the locations where the best $p$ responses are found. Maximum likelihood  estimation is then used to estimate the parameters based on the points in the initial design.
\end{itemize}

\subsection{User defined problem classes and/or policies}
Each of the problem classes is organized in its own .m file in the \code{./problemClasses} folder. The
standard API is defined as:\\
\code{function [mu, beta_W, numD]=UserDefinedName(varargin)}\\
     ~~~~ \code{          \%user defined function}\\
\code{end}\\
where varargin is used to pass input parameters with
variable lengths for the problem class if needed. \code{mu}  is a column vector generating a true function value (not known to the learner) of the user defined problem class. If Gaussian process regression is used, \code{mu} is sampled from the prior distribution. \code{beta_W} is a column vector of the inverse of the variance of measurement noise $\sigma^{W,2}_x$ and  \code{numD} specifies the dimensionality of the function. For example, if the user is intended to define their own GPR problems, the function should be defined as:
\begin{verbatim}
function [mu, beta_W, numD] = UserDefinedName( varargin )}
     mu_0=varargin{1};
     covM=varargin{2};
     beta_W=varargin{3};
     mu=mvnrnd(mu_0, covM)';
     numD=1;
end
\end{verbatim}
The user also needs to provide the priors for the GPR problems. This can be done by writing a \code{Prior_UserDefinedName.m} file in the \code{./Prior} folder to provide the prior mean, covariance matrix and the measurement noise:\\
\code{
function [mu_0, covM, beta_W] = Prior_UserDefinedName( varargin )
}

If the user would like to define a problem from a test function (such as Pinter's function with additive noise introduced in Section \ref{pc}), one should first  code up the test function, choose a discrete set of  alternatives and calculate their function values. One example can be:
\begin{verbatim}
function [mu, beta_W, numD] = Pinter( varargin )
     %set the dimensionality of the function 
     numD=2;
     
     %choose the discrete set of alternatives of interest 
     xx=-3:0.5:3;
     yy=-3:0.5:3;
     [x,y]=meshgrid(xx,yy);
     f=fun(x,y);
     [a,b]=size(f);
     mu=max(max(f))-reshape(f,[a*b,1]);
     
     %define the variance of measurement noise
     beta_W=1./(0.2*(max(max(f))-min(min(f)))).^2*ones(length(mu),1);
end

%define the test function
function f=fun(x,y)
      f=x.^2+2*y.^2+20*sin(y.*sin(x)-x+sin(y)).^2+40*sin(x.*sin(y)-y+sin(x)).^2
       +log10(1+(y.^2-2*x+3*y-cos(x)+1).^2) 
       +2*log10(1+2*(x.^2-2*y+3*x-cos(y)+1).^2)+1+rand()*5;
end
\end{verbatim}

Each of the policies is also organized in its own .m file in the ./policies folder. The standard
API is defined as:
\begin{verbatim}
function [mu_est, count, recommArm] = Name( mu_0,beta_W,covM,samples,alpha,tune) 
     [M,N]=size(samples);
     count=zeros(M,1);  % count the times each alternative is measured
     mu_est=mu_0;
     for i = 1:N 
          %user defined policy decision rule to find x=X^{\pi}(S^n)
          count(x)=count(x)+1;
          W=samples(x, count(x));
          %user defined transition function S^{n+1} = S^{M}(S^n, x,W)
     end
     [value, recommendedArm] = max(mu_est);
end
\end{verbatim}
where \code{mu_0} and \code{covM} 
specifies the prior distribution, \code{beta_W} is the known measurement (experimental) noise, \code{samples} are pre-generated
and shared among all the policies, \code{alpha} is the tunable parameter and \code{tune} specifies whether to tune
this policy or use default value.

\section{Experiments for Offline (Terminal Reward) Problems} \label{rs}
In this section we report on a series of experiments with the goal of illustrating the use of \pkg{MOLTE} and the types of reports that it produces.  We do not attempt to demonstrate that any policy is better than another, but our experiments support the hypothesis that different policies work well on different problem classes.  This observation supports the claim that more careful empirical work is needed to develop a better understanding of which policies work best, and under what conditions.

\subsection{Experiments with independent beliefs}
\begin{figure}
    \centering
   \subfigure[AUF:  Opportunity cost]{\includegraphics[width=0.43\textwidth]{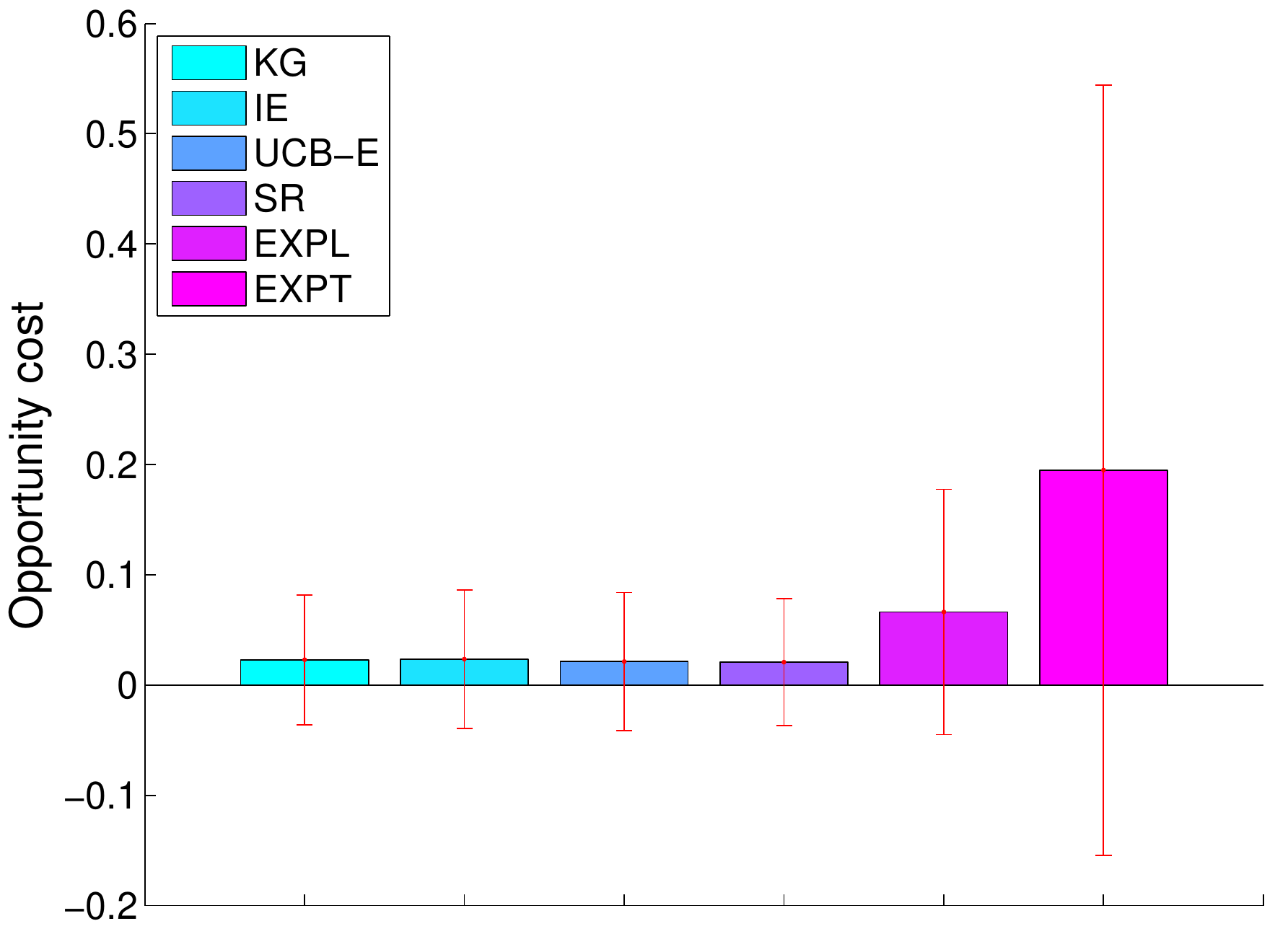}} ~~~
   \subfigure[AUF: Probability of optimality/winning]{   \includegraphics[width=0.435\textwidth]{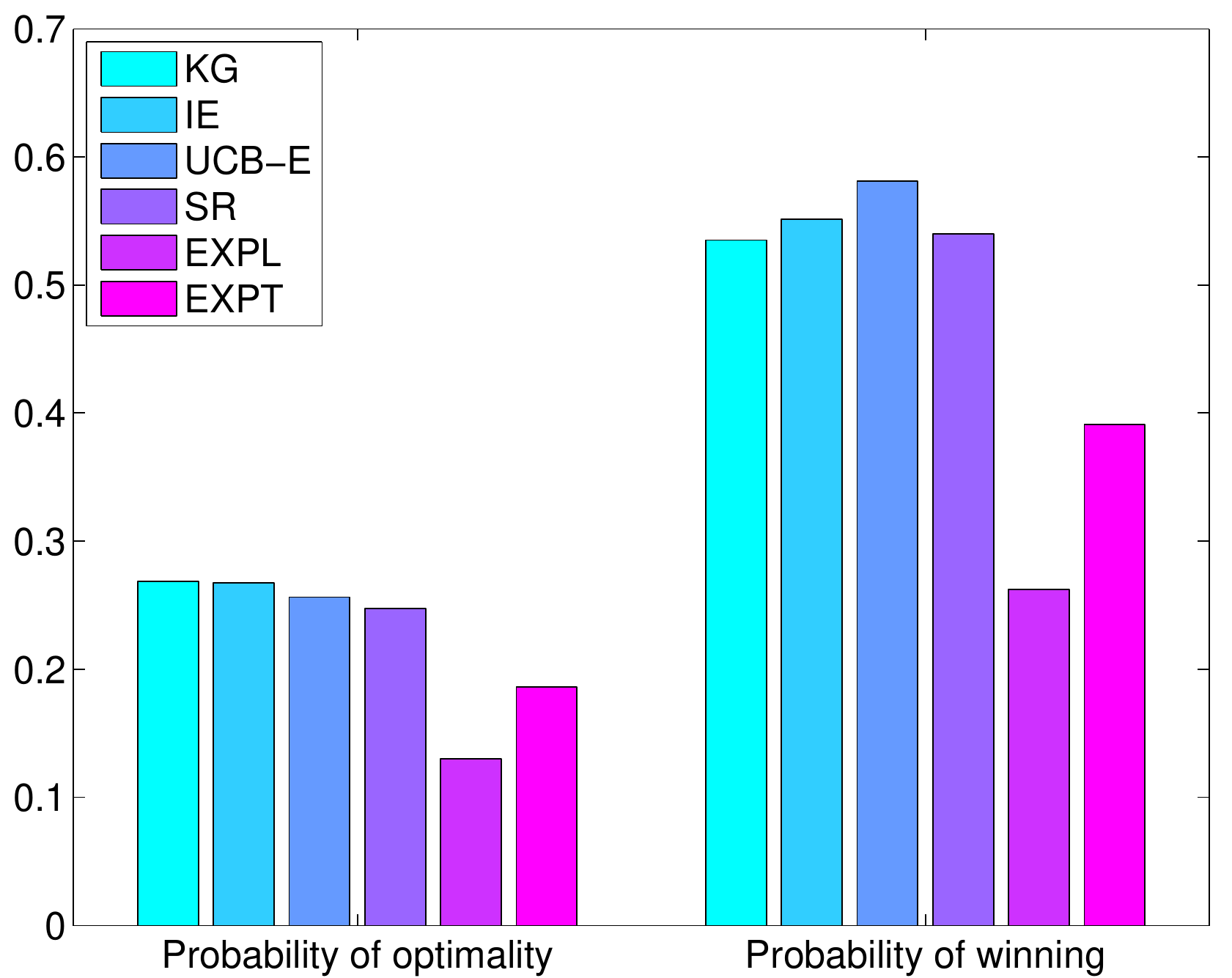}}\\
      \subfigure[Goldstein: Opportunity cost]{   \includegraphics[width=0.43\textwidth]{fig/OC_GOLD_scissored.pdf}}~~~
   \subfigure[Goldstein:Probability of optimality/winning]{\includegraphics[width=0.435\textwidth]{fig/Bar_Goldstein_scissored.pdf}}
    \caption{Comparisons for AUF and Goldstein. (a) and (c) depict the mean opportunity cost with error bars indicating the standard deviation of each policy. The first bar group in (b) and (d) demonstrates the probability that the final recommendation of each policy is the optimal one.  The second bar group in (b) and (d)  illustrates the probability that the opportunity cost of each policy is the lowest. }
    \label{bu}
\end{figure}

We first compare the performance of KG, IE with tuning, UCB-E with tuning, SR, EXPL and EXPT for offline ranking and selection problems. MLE is used to construct the prior distribution for KG and IE.
 Figure \ref{bu} shows the performance in problem classes AUF and Goldstein  with independent beliefs under a measurement budget  five times the number of alternatives.

We run each policy for \code{numP=1000} times.  We  illustrate in the first column of Figure \ref{bu} the mean opportunity cost and the standard deviation of each policy  over 1000 runs, with the opportunity cost (OC$^{\pi}$)  defined as: $$\text{OC}^{\pi}=\max_x \mu_x - \mu_{x^{\pi,N}},$$
where $x^{\pi,N}=\arg \max_{x}\theta^{\pi,N}_x$.

In order to provide a more comprehensive  comparison of different policies, we also calculate the probability that the final recommendation of each policy is the optimal one and   the probability that the opportunity cost of each policy is the lowest, as illustrated in the figures on the right hand side of Figure  \ref{bu}. 

The three criteria characterize the behavior of policies in different aspects. For example, under AUF, if one cares about the average performance of the policy and its stability, SR is the best choice concluding from Figure \ref{bu} (a). Yet, if one can only run one trial (as in most cases of experimental science) and want to identify the best alternative, KG might be a better choice since it has the highest probability of finding the optimal alternative. Or if one can live with fairly good alternatives other than the optimal one, UCB-E could be the choice (although it has to be carefully tuned).

One observation is that there is no universal best policy for all problem classes or under all criteria. In practice, a useful guidance could be abstracting the real world problem and running synthetic simulations to find the best simulated policy under some desired criterion   before conducting the real experiments.

 \subsection{Experiments with correlated beliefs}
In this section, we exploit correlated beliefs between alternatives in order to strengthen the effect of each measurement so that one measurement of some alternative can provide information for other alternatives.

  \begin{figure}[htp!]
\begin{tabular}{>{\centering\arraybackslash}m{.01\linewidth}MM}
\begin{turn}{90} KGCB \end{turn}&\includegraphics[width=0.48\textwidth]{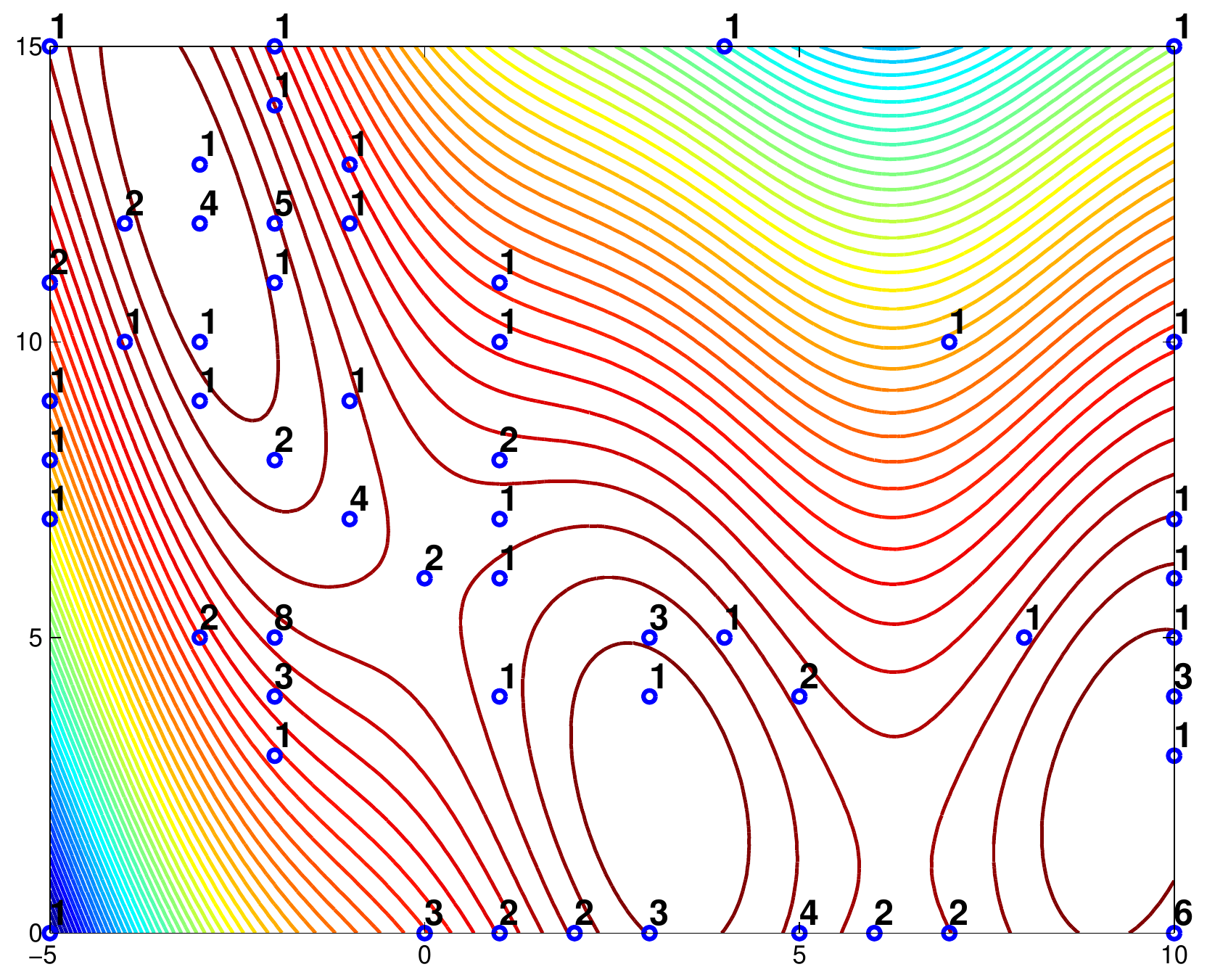}  &~\includegraphics[width=0.48\textwidth]{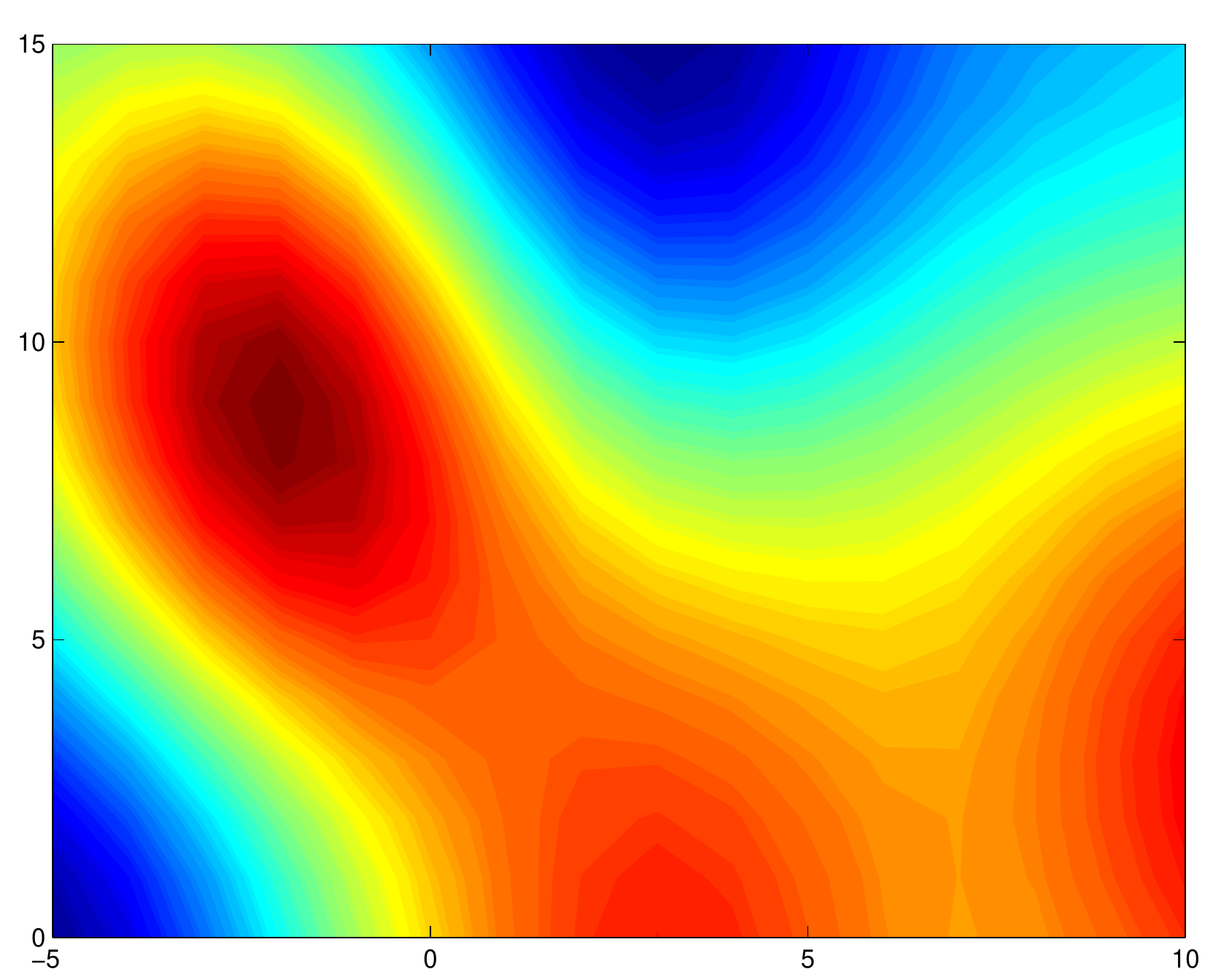} \\[-7.5pt]
\begin{turn}{90} Kriging \end{turn}&\includegraphics[width=0.48\textwidth]{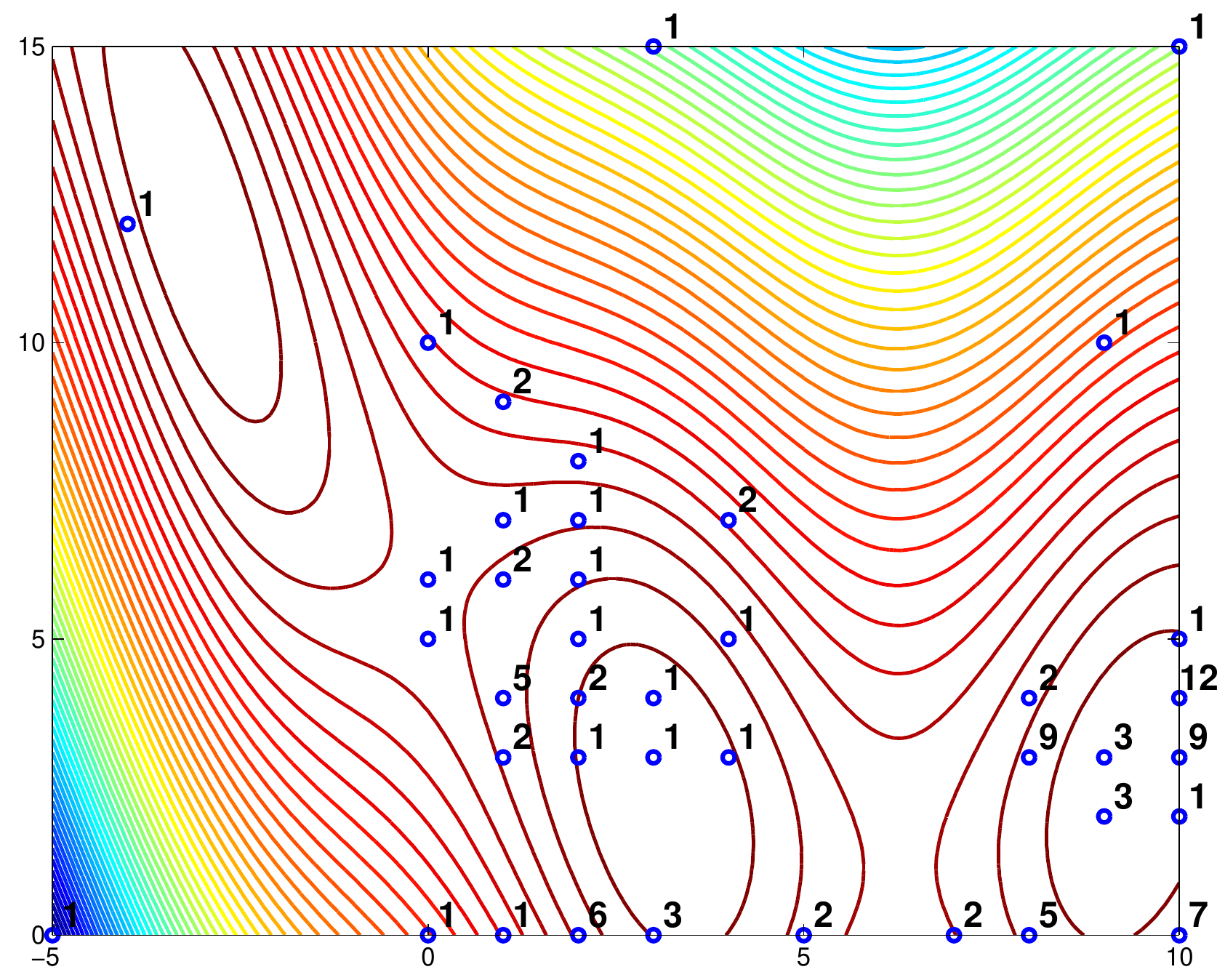}  &~\includegraphics[width=0.48\textwidth]{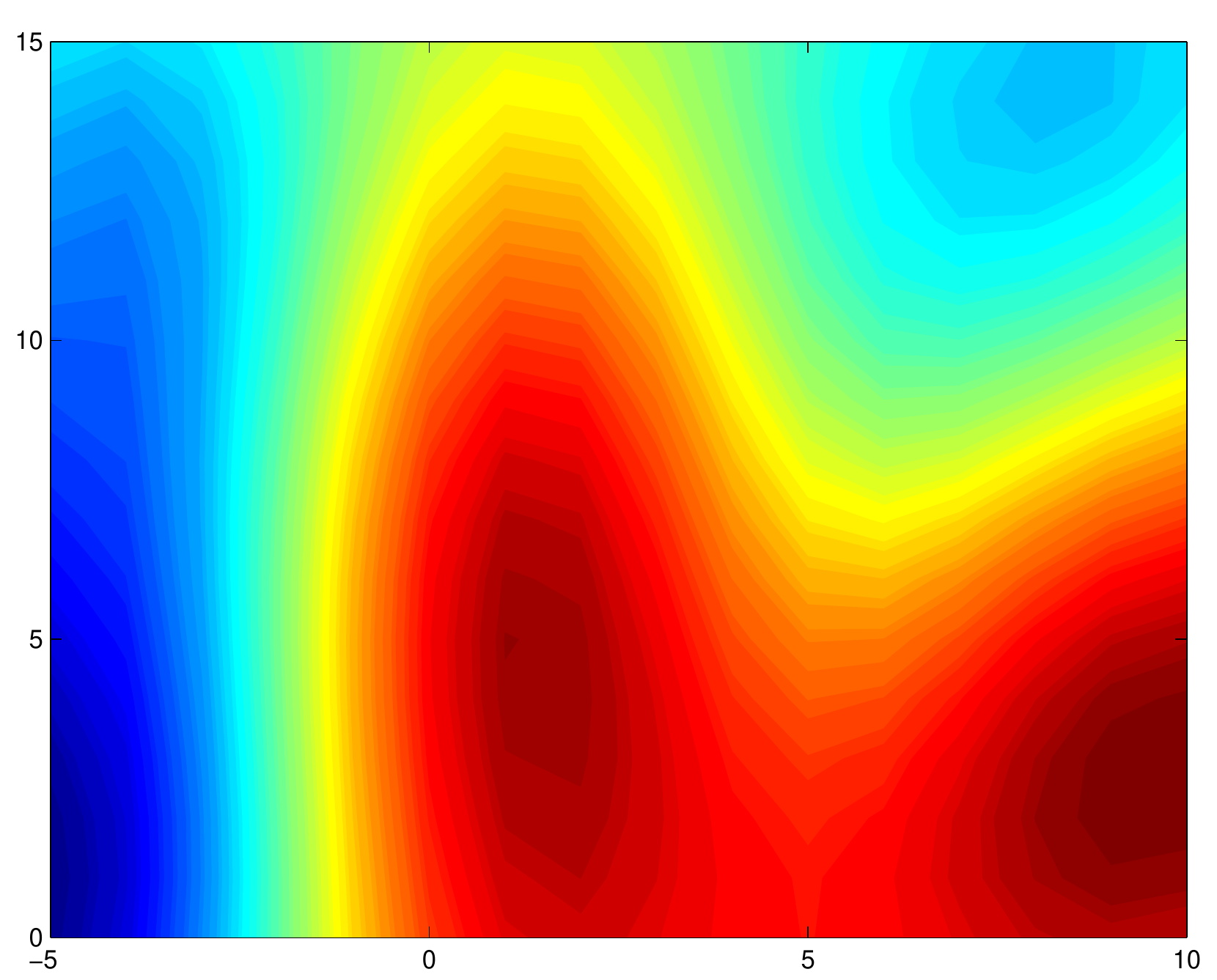} \\[-6pt]
\begin{turn}{90} UCB-E \end{turn}&\includegraphics[width=0.491\textwidth]{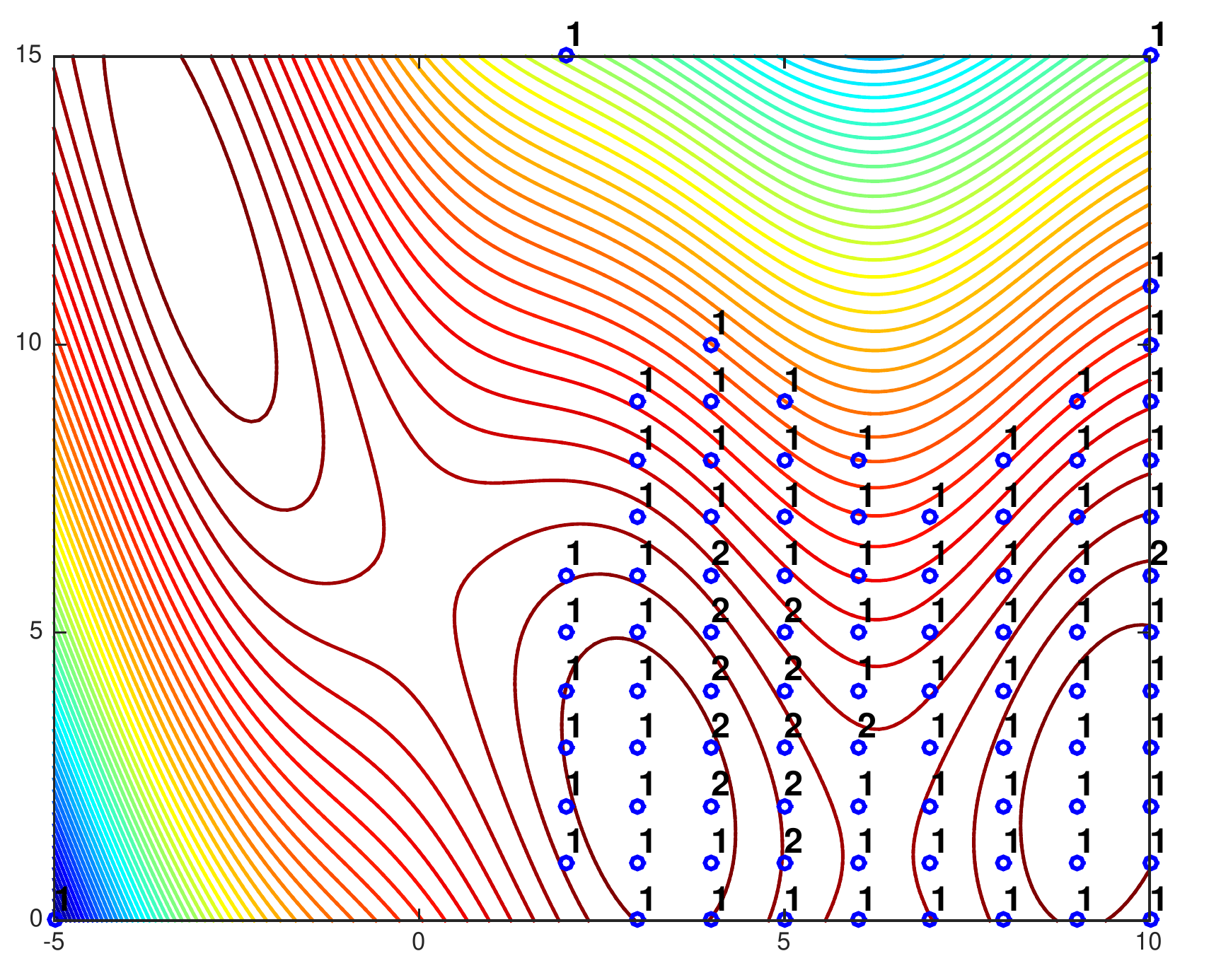}  &\includegraphics[width=0.499\textwidth]{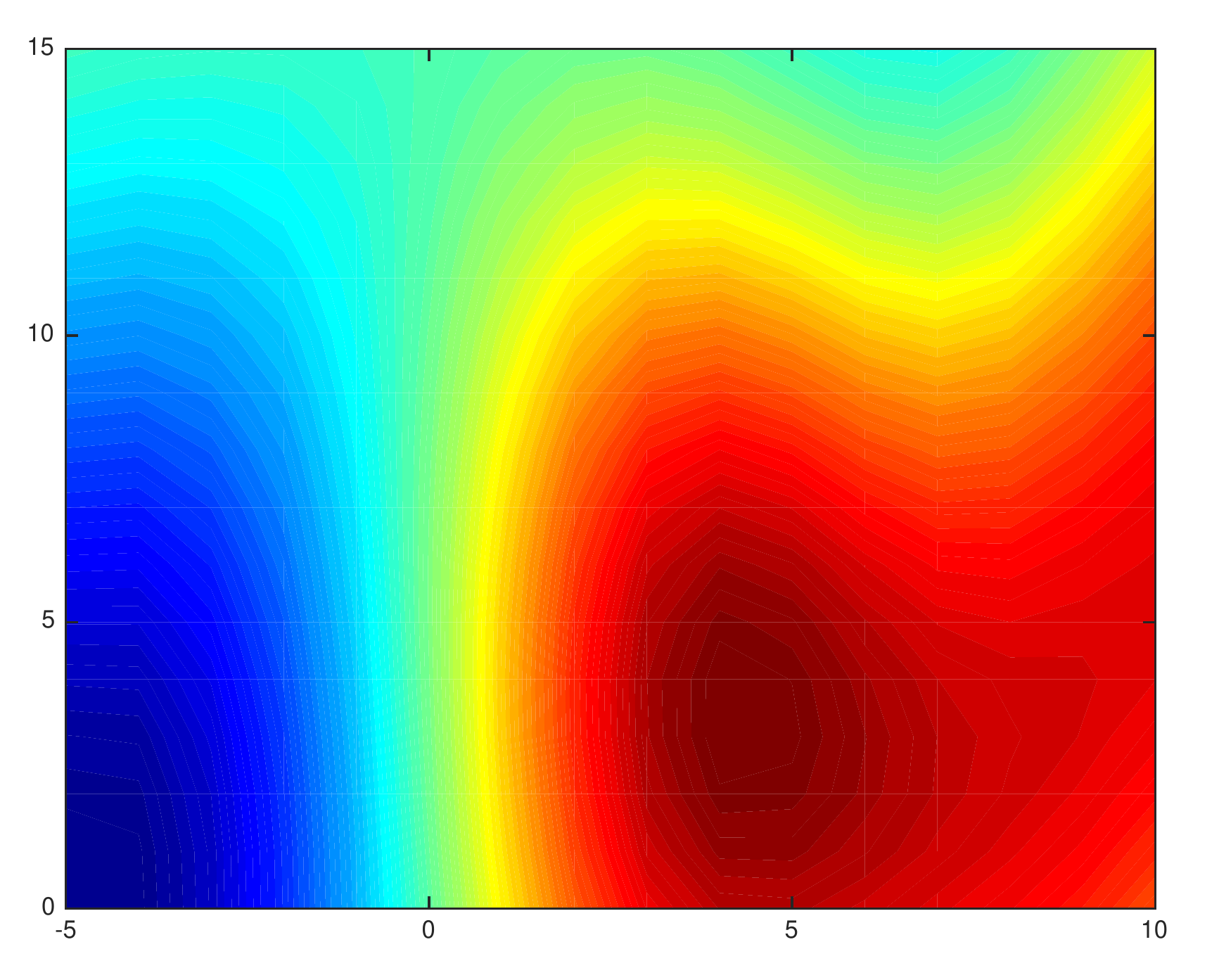} \\[-7.5pt]
\begin{turn}{90} Pure Exploitation \end{turn}&\includegraphics[width=0.483\textwidth]{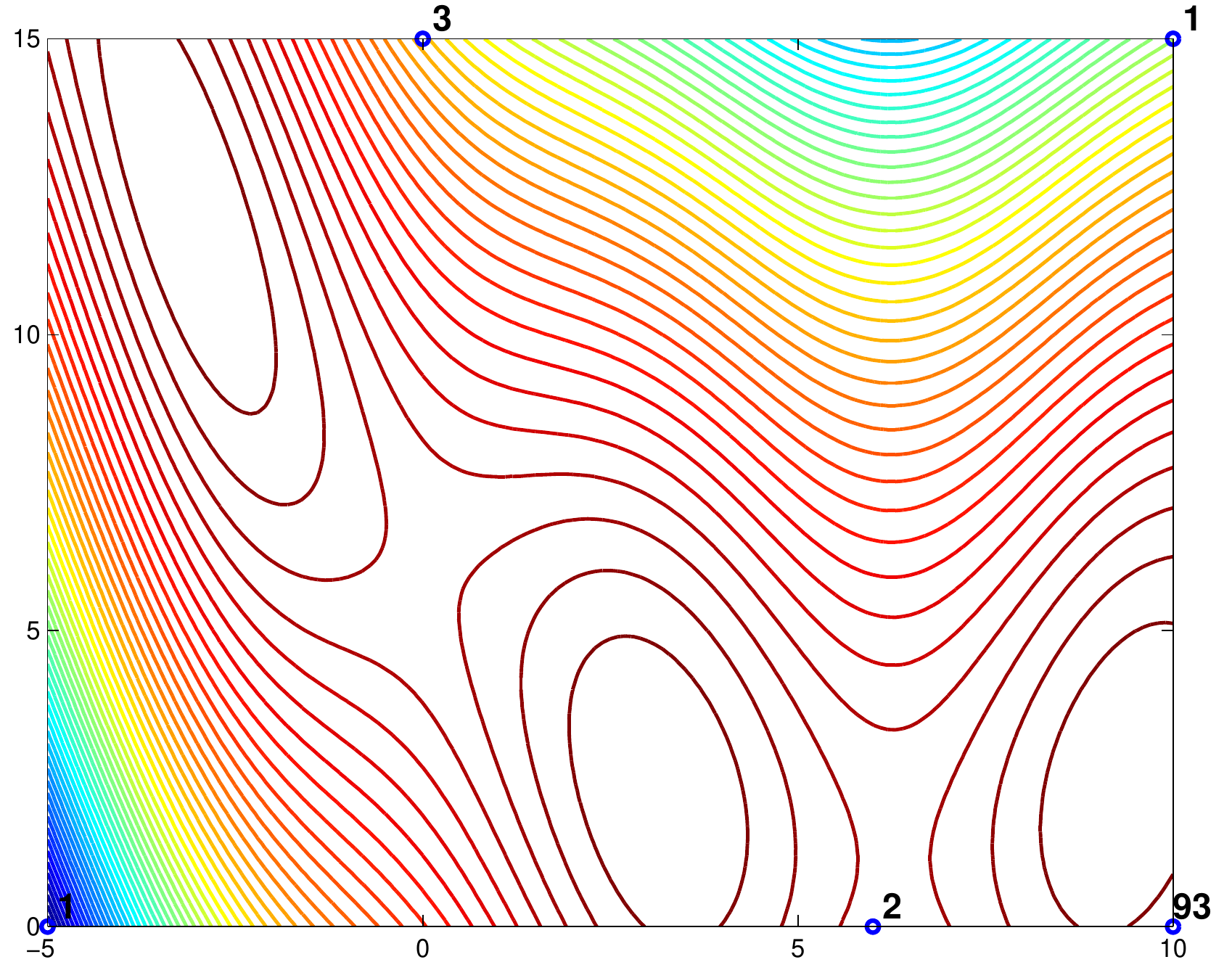}    &~\includegraphics[width=0.48\textwidth]{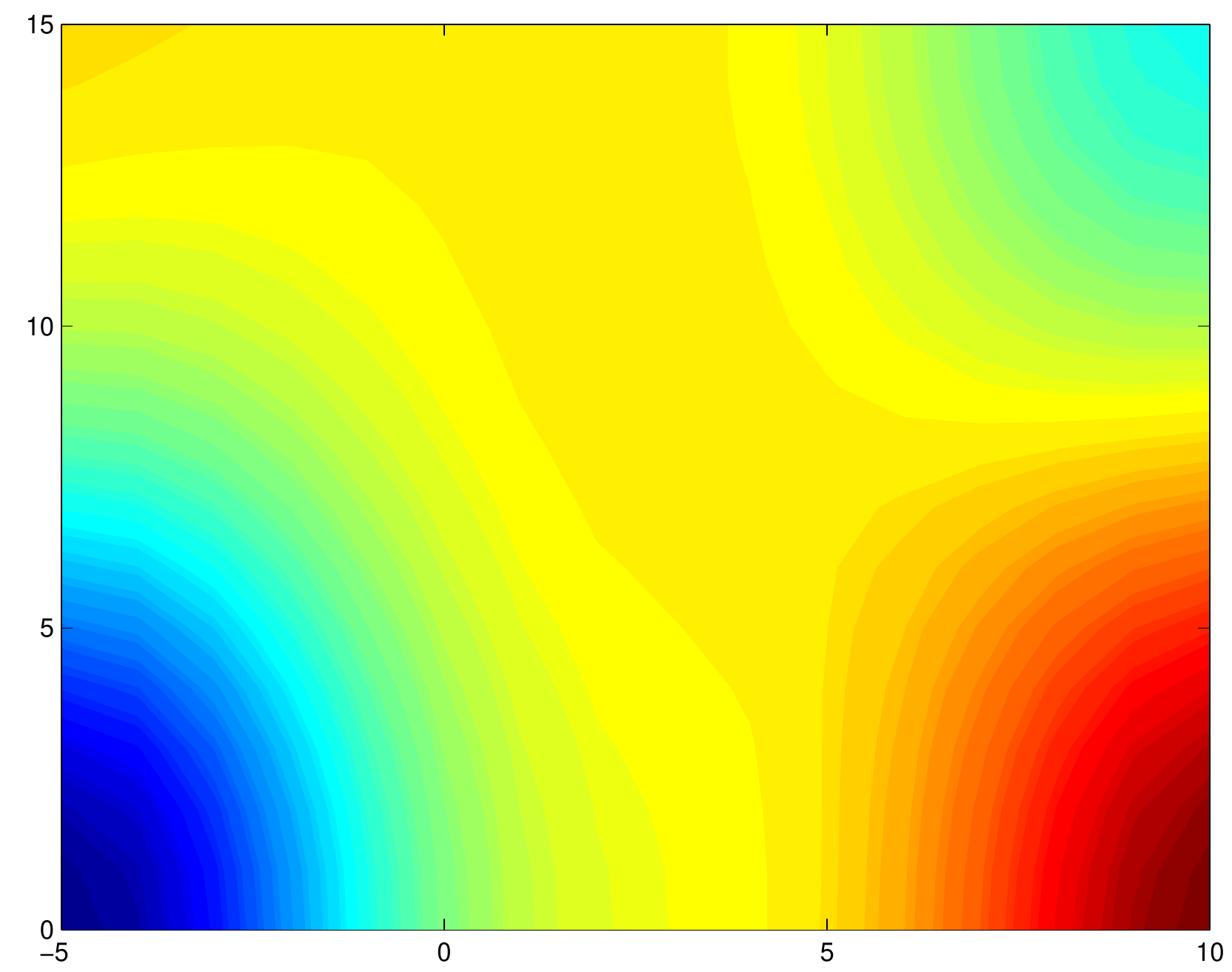}  \\[-8pt]
\end{tabular}
\caption{Left column: sampling distribution. Right column: posterior distribution.}
\label{lr}
\end{figure}

In order to better understand the behavior of each policy, a useful way is to examine the sampling pattern of each policy. We present  an example of the frequency of  measuring each alternative for each competing policy  for  Branin functions with a measurement budget of 100. To take advantage of correlated beliefs, rather than measuring each alternative once to initialize the empirical mean, we use the prior mean as the starting point and use the posterior mean $\theta^n$ in place of the empirical mean $\hat\theta^n$ for UCB-E. In the left column of Figure \ref{lr}, the sampling pattern of each policy is displayed together with the contour of the Branin objective function which exhibits one global maximum at $(-3, 12)$ and other two local maxima at $(9, 3)$ and $(16,4)$. The frequency that each alternative is measured is marked in numbers.  The right column depicts the final prediction under each policy. All the observations are pre-generated and shared for all policies. We see from the figures that since KGCB and Kriging take correlation into consideration in the decision functions, they need less exploration and rely on the correlation to provide information for less explored alternatives.  They quickly begin to focus on the alternatives that have the best values. Yet Kriging wanders around local minima for a while before it heads toward the global maximum. Note that the prediction of KGCB gives a good match in general. The function value at the true maximum alternative is well approximated, while moderate error in the estimate is located away from this region of interest. UCB-E is exploring more than necessary and wasting time on less promising regions. But when the budget is big enough, the exploration will contribute to better prediction of the surface, leading to a potentially larger final  outcome in the long run. Pure exploitation gets stuck in a seemingly good alternative and the sampling pattern is not reasonable nor meaningful.

\section{ Experiments  for Online (Cumulative Reward) Problems} \label{bandit}
In this section, we provide sample comparisons of different policies using the online (cumulative reward) objective function.
 The performance measure that we use to evaluate a policy $\pi$  in an online setting is  $\frac{\bar{R}^\pi_N}{N}$, where the pseudo-regret $\bar{R}_N^\pi$ is defined as
$$\bar{R}^\pi_N= N\max_{x \in \mathcal{X} }\mu_x - \sum_{n=0}^{N}\mathbb{E} [\mu_{X^{\pi, n}(s^n)}].$$ The opportunity cost (OC) between two policies in an online setting is defined as the difference of their pseudo-regrets.

\subsection{Experiments with independent beliefs}
In real world problems, especially in experimental science,  frequentist techniques cannot incorporate prior knowledge from domain experts, relying instead on the training from vast pools of data. This may be infeasible to perform in reality since running one experiment might be very expensive. The advantage of a Bayesian approach is unarguable in such cases. However, if we use MLE to fit the prior instead of using domain knowledge, it seems that the comparisons are in favor of Bayesian approaches by using an extra $11 \times p$ measurements. In order to make a seemingly more fair comparison in our synthetic experimental setting, we  also experiment with uninformative priors with no additional information provided for Bayesian approach.  

Tables \ref{10}, \ref{100} and \ref{500} provide comparisons of OLKG, IE with tuning,  UCB-E with tuning, UCB, UCB-V, KLUCB, pure exploration (EXPL)  using the Bubeck problems with uninformative prior. The measurement budgets are set to 10, 100 and 500 times the number of alternatives of each problem class in Tables \ref{10}, \ref{100} and \ref{500}, respectively. IE and UCB-E are carefully tuned for each problem class. Under each problem class, we ran each policy for \code{numP=1000} times. In each run, all the measurements are pre-generated and shared across all the policies.  For each policy we record the normalized opportunity cost between OLKG and other competing policies, where the normalized opportunity cost is defined as  the ratio between the opportunity cost $\frac{\bar{R}^\pi_N}{N}-\frac{\bar{R}^\text{OLKG}_N}{N} $ and the range of the truth $\mu$. Positive values of OC indicate that the corresponding policy underperforms OLKG on average.   Other than the  interest of average performance measured by pseudo-regret, only one sample path will be realized in real world experiments and  it is meaningful to find out which policy is most likely to perform the best in one sample run.  Thus we also report  the probability that  each of the other policy outperforms (obtains a lower regret than) OLKG within 1000 realizations.  Any policy can be set as a benchmark by placing it as the first policy in the input spreadsheet.  

We see from the three tables that  the probability of any other policy that outperforms OLKG is in general much less than 0.5. If this criterion is what an experimenter anticipates, then OLKG is a safe choice in most situations.  We then discuss the performance of each policy in terms of OC.  At the beginning of each trial,  IE and UCB-E are more exploiting than exploring while OLKG tends to explore before it moves toward the best estimates.  This contributes to  good performance (measured by OC) of IE and UCB-E in Table \ref{10} with a small measurement budget.    The tuned values of parameters further sharpen this effect by utilizing  smaller values  compared to those under larger measurement budgets as reported in Table \ref{tuningV} which summarizes the optimally tuned values for each parameter.  Since UCB policies tend to explore more than necessary (which can be seen from the sampling pattern, for example,  Figure \ref{lr}), the performance degenerates with a moderate measurement budget as shown in Table \ref{100}. In this case, OLKG yields the best performance since after an exploration period, it begins to focus on the alternatives that  have the best estimates while  looking for alternatives whose estimates are less certain. Yet exploration benefits in the long run. Thus the performance of UCB policies and IE improves if allowed to explore for a sufficiently long time as reported in Table \ref{500}. 

\begin{table}
\centering
\setlength{\tabcolsep}{1.8pt}
\caption{The difference between each policy and OLKG (OC), and the probability that each policy outperforms OLKG, using uninformative priors with a measurement budget 10 times the number of alternatives.}
\begin{tabular}{|c|cc|cc|cc|cc|cc|cc|cc|cc|cc|cc|cc|cc|cc|cc|}
    \hline
 \multirow{2}{*}{Problem Class}    &\multicolumn{2}{c|}{IE}    &\multicolumn{2}{c|}{UCBE}    &\multicolumn{2}{c|}{UCBV}  &\multicolumn{2}{c|}{UCB} &\multicolumn{2}{c|}{KLUCB} &\multicolumn{2}{c|}{EXPL}\\
\cline{2-13}
&OC & Prob. &OC & Prob. &OC & Prob. &OC & Prob. &OC & Prob.  &OC & Prob.  \\
   \hline
Bubeck1  & -0.031 & 0.43 & -0.032  & 0.43 & 0.073  & 0.51   & 0.016 & 0.35   & 0.054  & 0.50  & 0.078  & 0.50 \\ 
Bubeck2   & -0.032 & 0.55 & -0.031  & 0.52 & 0.097   & 0.30 & 0.025  & 0.43  & 0.070  & 0.35  & 0.105  & 0.29 \\ 
Bubeck3 & -0.000  & 0.29 & 0.006  & 0.30 & 0.068    & 0.26  & 0.021  & 0.53  & 0.020  & 0.34  & 0.095  & 0.23 \\ 
Bubeck4 & -0.004  & 0.39 & -0.003  & 0.57 & 0.100   & 0.36  & 0.029  & 0.48  & 0.040  & 0.40  & 0.124  & 0.33 \\ 
Bubeck5  & -0.019  & 0.71 & -0.020  & 0.71 & 0.213 & 0.01  & 0.018  & 0.48  & 0.087  & 0.11  & 0.255  & 0.00 \\ 
Bubeck6 & -0.034  & 0.49 & -0.035  & 0.48 & 0.139  & 0.34  & 0.034  & 0.41  & 0.098  & 0.37  & 0.151  & 0.33 \\ 
Bubeck7  & -0.036  & 0.70 & -0.036  & 0.71 & 0.065 & 0.17  & 0.009  & 0.48  & 0.043  & 0.22  & 0.073  & 0.15 \\ \hline
\end{tabular}
\label{10}
\end{table}

\begin{table}
\centering
\setlength{\tabcolsep}{1.8pt}
\caption{The difference between each policy and OLKG (OC), and the probability that each policy outperforms OLKG, using uninformative priors with a measurement budget 100 times the number of alternatives.}
\begin{tabular}{|c|cc|cc|cc|cc|cc|cc|cc|cc|cc|cc|cc|cc|cc|cc|}
    \hline
 \multirow{2}{*}{Problem Class}    &\multicolumn{2}{c|}{IE}    &\multicolumn{2}{c|}{UCBE}    &\multicolumn{2}{c|}{UCBV}  &\multicolumn{2}{c|}{UCB} &\multicolumn{2}{c|}{KLUCB} &\multicolumn{2}{c|}{EXPL}\\
\cline{2-13}
&OC & Prob. &OC & Prob. &OC & Prob. &OC & Prob. &OC & Prob.  &OC & Prob.  \\
   \hline
Bubeck1 & 0.006  & 0.34   & 0.015  & 0.32   & 0.387 & 0.36   & 0.245  & 0.14  & 0.311  & 0.37 & 0.431  & 0.36 \\ 
Bubeck2 & 0.006 & 0.31    & 0.017 & 0.35    & 0.399 & 0.09   & 0.226  & 0.17 & 0.309 & 0.22 & 0.458  & 0.06 \\ 
Bubeck3 & 0.002  & 0.32   & 0.007  & 0.31   & 0.111  & 0.18  & 0.077  & 0.39 & 0.052 & 0.25 & 0.214  & 0.07 \\ 
Bubeck4 & -0.014 & 0.31   & -0.005 & 0.30   & 0.232  & 0.27  & 0.156  & 0.32 & 0.114  & 0.30 & 0.365 & 0.17 \\ 
Bubeck5 & -0.003 & 0.39   & 0.003  & 0.34   & 0.228 & 0.01   & 0.064  & 0.26 & 0.094  & 0.15 & 0.425 & 0.00 \\ 
Bubeck6 & 0.014  & 0.38   & 0.025  & 0.38   & 0.522 & 0.10   & 0.274  & 0.12 & 0.380  & 0.10 & 0.619  & 0.09 \\ 
Bubeck7 & 0.015 & 0.52    & 0.016 & 0.44    & 0.260 & 0.00 & 0.158  & 0.21 & 0.215  & 0.09 & 0.303  & 0.00 \\  \hline
\end{tabular}
\label{100}
\end{table}

\begin{table}
\centering
\setlength{\tabcolsep}{1.8pt}
\caption{The difference between each policy and OLKG (OC), and the probability that each policy outperforms OLKG, using uninformative priors with a measurement budget 500 times the number of alternatives.}
\begin{tabular}{|c|cc|cc|cc|cc|cc|cc|cc|cc|cc|cc|cc|cc|cc|cc|}
    \hline
 \multirow{2}{*}{Problem Class}    &\multicolumn{2}{c|}{IE}    &\multicolumn{2}{c|}{UCBE}    &\multicolumn{2}{c|}{UCBV}  &\multicolumn{2}{c|}{UCB} &\multicolumn{2}{c|}{KLUCB} &\multicolumn{2}{c|}{EXPL}\\
\cline{2-13}
&OC & Prob. &OC & Prob. &OC & Prob. &OC & Prob. &OC & Prob.  &OC & Prob.  \\
   \hline
Bubeck1  & -0.105  & 0.30 & -0.098  & 0.30  & 0.296  & 0.26   & 0.288  & 0.10   & 0.175 & 0.27   & 0.634  & 0.26 \\ 
Bubeck2 & -0.089   & 0.28 & -0.080 & 0.26   & 0.253 & 0.31    & 0.226  & 0.15   & 0.139  & 0.32 & 0.609 & 0.02 \\ 
Bubeck3  & -0.009  & 0.34 & -0.006  & 0.31  & 0.069  & 0.18   & 0.077  & 0.39   & 0.035  & 0.29 & 0.268  & 0.03\\ 
Bubeck4 & -0.075   & 0.28 & -0.069 & 0.27   & 0.091 & 0.26    & 0.174  & 0.24   & 0.014  & 0.26 & 0.462  & 0.12 \\ 
Bubeck5  & -0.030 & 0.33 & -0.026  & 0.31   & 0.066  & 0.28   & 0.050 & 0.23    & 0.012  & 0.34 & 0.462  & 0.00 \\ 
Bubeck6  & -0.024  & 0.26 & -0.022  & 0.24  & 0.310 & 0.05    & 0.227 & 0.16    & 0.190  & 0.06 & 0.771  & 0.05 \\ 
Bubeck7  & -0.045   & 0.33 & -0.045  & 0.34 & 0.262 & 0.11    & 0.152 & 0.23    & 0.200  & 0.27 & 0.430 & 0.00 \\ 
 \hline
\end{tabular}
\label{500}
\end{table}

\begin{table}
\centering
\caption{Tuned parameters of IE and UCB-E under different problem classes and measurement budgets. The second row indicates the ratio between the measurement budget and the number of alternatives. }
\begin{tabular}{|c|ccc|ccc|} \hline
\multirow{2}{*}{Problem Class}    &\multicolumn{3}{c|}{IE}    		&\multicolumn{3}{c|}{UCBE}  \\
\cline{2-7}
	    					&10 		& 100		&500    		&10 		& 100		&500  \\ \hline
	Bubeck1 		&0.0007079		&1.295     &	2.036		&0.0008991			&0.3934			&1.103	\\	  		
	Bubeck2			&0.1675			&1.295 &2.169 			&0.002359			&0.337			&0.9063	\\
	Bubeck3 		   	&0.8991		&1.395       &1.878			&0.1206		&0.4562			&0.8635	 \\
	Bubeck4 				&0.8991		&1.571    & 2.196               & 0.004392     & 0.5332   &1.197\\
	Bubeck5				&0.004566			&1.395&2.169          &0.0003102        &0.3518    &1.002\\
	Bubeck6 			&0.09063			&1.197     & 1.642             &  0.000505     &0.3201     &0.7748\\
	Bubeck7		&0.002773			&0.8991             &1.878             & 0.0005936  &0.2169     &0.8007\\ \hline
\end{tabular}
\label{tuningV}
\end{table}

\subsection{Experiments with correlated beliefs}
In this section, we summarize numerical experiments on problems with correlated beliefs between different policies, including OLKG, IE with tuning, UCBE, UCB-V,  Kriging, UCB, Thompson Sampling (TS) and pure exploration (EXPL). 
To take advantage of correlated beliefs, we use the prior mean as the starting point and use posterior mean $\theta^n$ in place of the empirical mean for UCB-V and UCB policies.

\begin{figure*}
    \centering
    \subfigure[Goldstein]{
        \label{fig_ol_100_1000}
        \includegraphics[width=0.31\textwidth]{fig/Goldstein_scissored.pdf}
    }
    \subfigure[HyperEllipsoid]{
        \label{fig_ol_100_150}
       \includegraphics[width=0.31\textwidth]{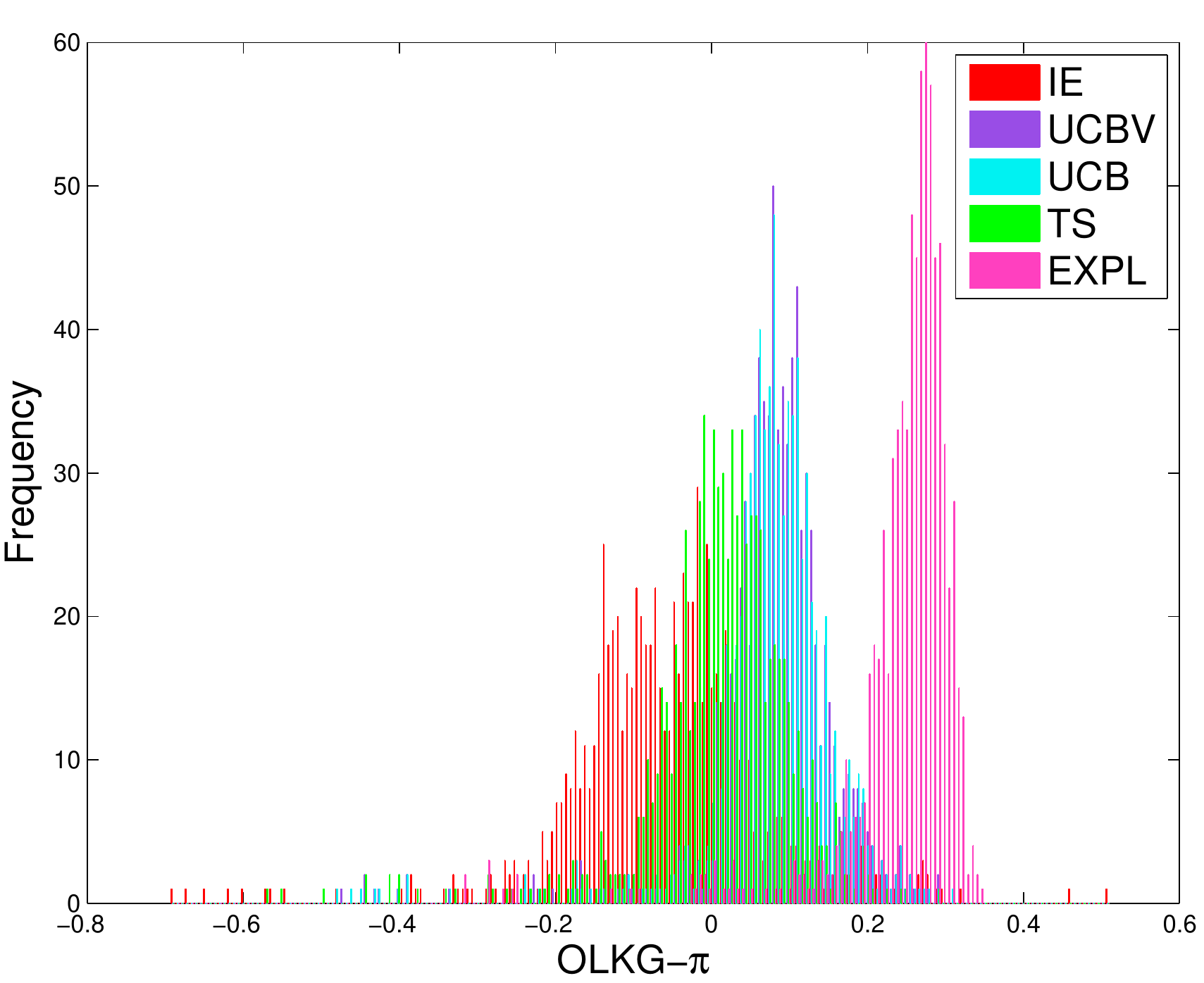}
    }
    \subfigure[Rastrigin]{
        \label{fig_ol_corr_100_30}
        \includegraphics[width=0.31\textwidth]{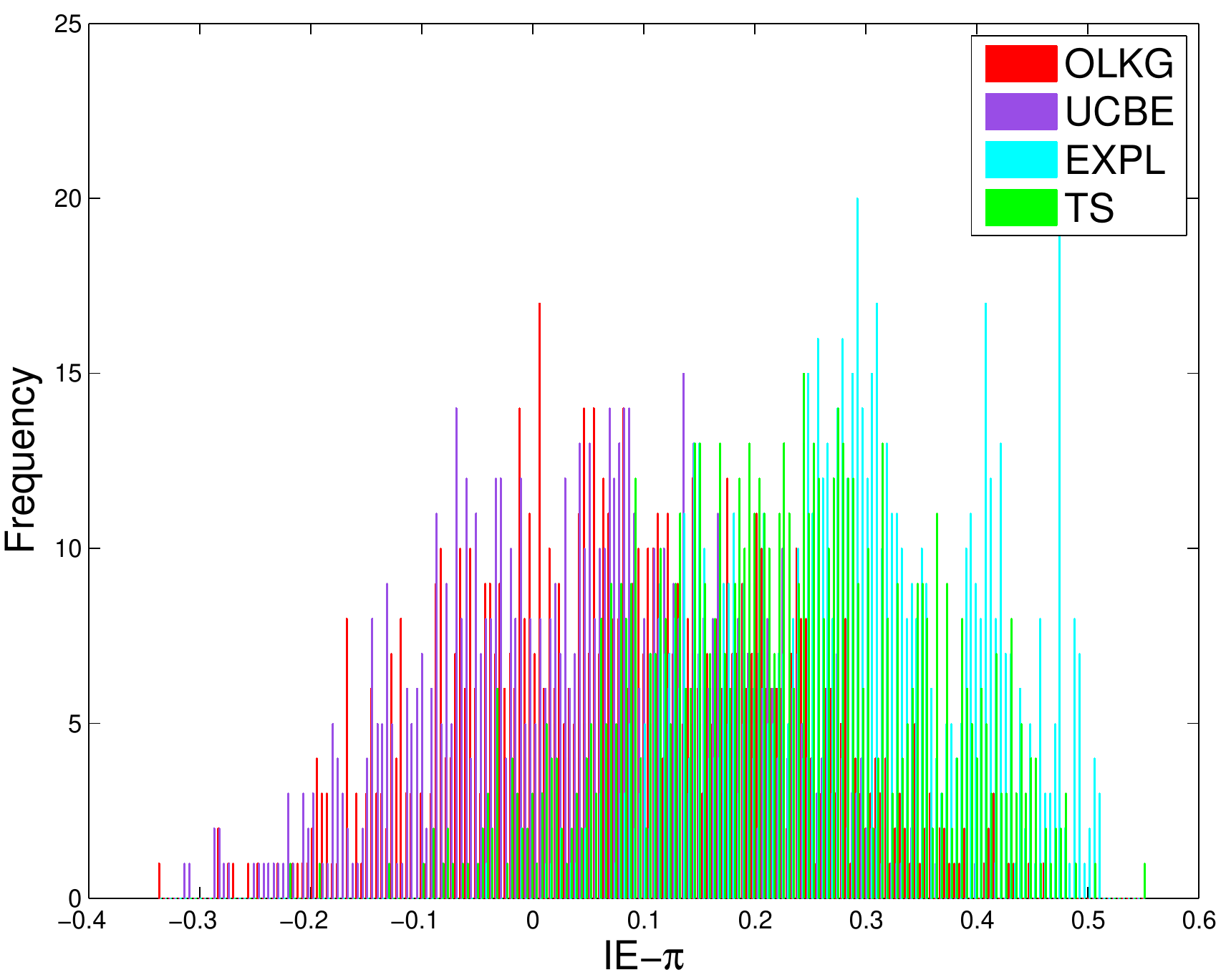}
    }

    \caption{Normalized opportunity cost between different policies.}
\label{histo}
\end{figure*}

In order to gain a good understanding of the performance of the policies, MOLTE produces histograms illustrating the distribution of the difference between the normalized OC of a benchmark policy and either of the other policies over 1000 runs. Whichever policy that is listed as the first policy is treated as the benchmark.  The measurement budget is set to $0.2$ times the number of alternatives of each problem class. Figure \ref{histo} compares the performance of several policies under various problem classes with different benchmark policies. A distribution centered around a positive value implies the policy underperforms the benchmark policy, while one centered around a negative number means the policy outperforms the benchmark. For example, Figure  \ref{fig_ol_100_1000} compares the performance of UCBV, OLKG, IE, TS and EXPL under Goldstein with UCBV as the benchmark policy. We can see that the tuned IE and OLKG are outperforming UCBV and others are underperforming. 

We close this section by providing  more comparisons between other policies with OLKG under various problem classes. The measurement budget is set to $0.2$ times the number of alternatives of each problem class. Table \ref{onlincom} reports the normalized mean OCs and the probability that each of the other policy outperforms OLKG under 1000 runs. IE and UCB-E are carefully tuned for each problem classes with the optimal value shown in Table \ref{TV}. IE and UCB-E after tuning works generally well. Yet the optimal values of the tuned parameters are quite different for different problems as shown in Table \ref{tuningV} and  \ref{TV}. In addition,  the performance of the policies are sensitive to the value of the tunable parameters.  In light of this issue, we can conclude that
OLKG and Kriging have one attractive advantage over IE and UCB-E: they require no tuning at all, while yielding comparable performance to a
finely tuned IE or UCB-E policy. A detailed study on the issue of tuning is presented in Section \ref{IEtuning}.

\begin{table}
\centering
\setlength{\tabcolsep}{1.8pt}
\caption{Comparisons with OLKG for correlated beliefs with the measurement 0.2 times the number of alternatives of each problem class.}
\begin{tabular}{|c|cc|cc|cc|cc|cc|cc|cc|cc|cc|cc|cc|cc|cc|cc|}
    \hline
 \multirow{2}{*}{Problem Class}    &\multicolumn{2}{c|}{IE}    &\multicolumn{2}{c|}{UCBE}  &\multicolumn{2}{c|}{UCBV}   &\multicolumn{2}{c|}{Kriging}  &\multicolumn{2}{c|}{TS} &\multicolumn{2}{c|}{EXPL}\\
\cline{2-13}
&OC & Prob. &OC & Prob. &OC & Prob. &OC & Prob. &OC & Prob.  &OC & Prob.  \\
   \hline
Goldstein & -0.061  & 0.81  		&-0.097 &0.92  & -0.003 & 0.45 	     & -0.031 & 0.73      & 0.100  & 0.09     & 0.041  & 0.16 \\ 
AUF\_HNoise & 0.058 & 0.40  		& 0.022 &0.43  & 0.037  & 0.54 	     & 0.031  & 0.39      & 0.073  & 0.22   	& 0.047  & 0.48 \\ 
AUF\_MNoise  & 0.043  & 0.29  	&0.027  &0.42	& 0.343  & 0.21		     & 0.023  & 0.28     & 0.173  & 0.21 	& -0.057  & 0.52 \\ 
AUF\_LNoise  & -0.043 & 0.73  		&-0.013 &0.64 & 0.053  & 0.51 	      & 0.005  & 0.53      & 0.038  & 0.20 	& 0.003 & 0.62      \\ 
Branin  & -0.027  & 0.76 		& 0.025  &0.24   & 0.026 & 0.26           & 0.004  & 0.54    & 0.041  & 0.07 	& 0.123  & 0.00 \\ 
Ackley  & 0.007  & 0.42  		&0.04  &0.41	& 0.106  & 0.20 	     & 0.037  & 0.42    & 0.100  & 0.23 	& 0.344 & 0.00 \\ 
HyperEllipsoid & -0.059  & 0.73 &0.064 &0.12 	&0.08 & 0.07        & 0.146  & 0.22      & 0.011 & 0.38 	& 0.243  & 0.03 \\ 
Pinter  & -0.028  & 0.56  	& -0.003  &0.51		& 0.029  & 0.42 	     & -0.055  & 0.65   & 0.122  & 0.19 	& 0.177  & 0.04 \\ 
Rastrigin  & -0.082  & 0.70	&-0.03 &0.56		 & 0.162  & 0.04 	   & -0.026  & 0.57    & 0.136 & 0.08 	    & 0.203  & 0.01 \\ 
 \hline
\end{tabular}
\label{onlincom}
\end{table}

Table \ref{onlincom} together with the comparisons shown in previous sections suggests that there is no universal best policy for all problem classes and  one could possibly design toy problems for either policy to perform the best.  Similar observations  have also been reported by  \cite{kuleshov2014algorithms} for different bandit problems on different metrics. Besides, there are theoretical guarantees proved for each of the policy mentioned above, but the existence of these bounds does not appear to provide reliable guidance regarding which policy works best. An asymptotic bound does not provide any assurance that an algorithm will work well on a particular problem in finite time. In practice, we believe that more useful guidance could be obtained by abstracting a real world problem, running simulations and using these to indicate which policy works best. 
\begin{table}
\centering
\caption{Tuned parameters of IE and UCB-E under different problem classes.}
\begin{tabular}{|c|cc|}\hline
Problem Class    & IE  &UCBE\\ \hline
Goldstein  	     &0.009939  &2571 \\
AUF\_HNoise     &  0.01497            &0.319\\
AUF\_MNoise   &0.01871 & 1.591 \\
AUF\_LNoise    & 0.01095 &6.835\\
Branin  &  0.2694 &0.0003664\\
Ackley  & 1.197   & 1.329\\
HyperEllipsoid &0.8991 &21.21\\
Pinter &0.9989  &0.0001636 \\
Rastrigin&0.2086  &0.001476 \\ \hline
\end{tabular}
\label{TV}
\end{table}

\section{Discussion}\label{Dis}
We close our presentation by discussing two issues that tend to be overlooked in comparisons of learning algorithms: the tuning of heuristic parameters (widely used in frequentist UCB policies) and priors (used in all Bayesian policies such as knowledge gradient).
\subsection{The issue of tuning}\label{IEtuning}
Previous experimental results show that tuned version of IE and UCB-E yield good performance in general and yet the optimal value for IE and UCB-E may be highly problem dependent. Our experiments also suggest that the performance of a policy is sensitive to the value of the tuned parameter. For example, Figure \ref{tuningS} provides the comparisons between the performances of IE with different parameter values (provided in the parentheses) with the online objective function under various problem classes. The measurement budget is set to five times the number of alternatives for each problem class experimented with independent beliefs and 0.3 times the number of alternatives for each problem class experimented with correlated beliefs. `OC' is the mean opportunity cost comparing tuned IE with others $OC^\text{IE}-OC^\pi$, with  a positive value indicating a win for tuned IE. `Prob.'  is the probability that other policies outperform the tuned IE. We see from the table that $z_\alpha$  is highly problem dependent and the performance degrades quickly away from the optimal value. For some experimental applications, tuning can require running physical experiments, which may be very expensive or even entirely infeasible.
\begin{table}[htp!].
\centering
\setlength{\tabcolsep}{3pt}
\begin{tabular}{|c|c|c|cc|cc|cc|cc|cc|cc}\hline
 \multirow{2}{*}{Problem Class}    & \multirow{2}{*}{B} &\multirow{2}{*}{$z_\alpha^*$ } &\multicolumn{2}{c|}{IE(1)}    &\multicolumn{2}{c|}{IE(2)}    &\multicolumn{2}{c|}{IE(3)}  &\multicolumn{2}{c|}{IE(4)} &\multicolumn{2}{c|}{IE(5)}\\
\cline{4-13}
& &  &OC & Prob. &OC & Prob. &OC & Prob. &OC & Prob. &OC & Prob.   \\ \hline
Bubeck4  & I &2.086 & 0.002 &0.40 & 0.001 & 0.45            &0.002 & 0.46        & 0.015 &0.47  & 0.017 & 0.47 \\
Bubeck6 & I  & 2.01 & 0.003 & 0.44 & 0.001 & 0.48             & 0.004 & 0.43      & 0.013 & 0.23 & 0.028  & 0.13\\ 
AUF\_MNoise & I  & 1.1305 & 0.004  & 0.38 & 0.041 & 0.04  & 0.071 & 0.00    & 0.095 & 0.00 & 0.114 & 0.09\\ 
CamelBack & I  & 1.295 & 0.006 & 0.35 & 0.069  & 0.32    & 0.108 & 0.03          & 0.145 & 0.00 & 0.172 & 0.00\\ 
AUF\_LNoise & C & 0.9498  & 0.043  & 0.00 & 0.080 & 0.00   & 0.105 & 0.00  & 0.123 & 0.03 & 0.136 & 0.00\\ 
Branin & C &0.4438   & 0.001  & 0.25 & 0.005 & 0.32          & 0.014 & 0.07           & 0.023  & 0.01 & 0.032  & 0.01\\ 
Goldstein & C & 0.079  & 0.071 & 0.00 & 0.090& 0.00       & 0.101  & 0.00          & 0.108 & 0.00 & 0.113 & 0.00\\ 
Rosenbrock & C & 0.9989   & 0.007  & 0.18 & 0.060 & 0.08  & 0.093 & 0.05        & 0.120 & 0.04 & 0.143 & 0.03\\ \hline
\end{tabular}
\caption{Comparisons between  tuned IE and IEs with fixed parameter values. The second column indicates the belief model, with I for independent belief and C  for correlated belief.  $z_\alpha^*$ is the tuned value for each problem class. The number included in the parenthesis is the parameter value used by each IE policy.}
\label{tuningS}
\end{table}

\subsection{The issue of constructing priors}
In MOLTE, we use MLE to fit the prior for test functions based on sampling measurements, which seems like a tuning process. Yet designing a Bayesian prior is not necessarily the same as  tuning parameters.   In real world problems, such as applications in experimental sciences (although there are many other examples from other problem domains), the Bayesian prior  may be based on an understanding of the physical system and might be based on the underlying chemistry/physics of  the problem, a review of the literature, or past experience.  This information might be qualitative in nature and is not easily incorporated by frequentist approaches.   When this domain knowledgeable is available, and especially when experiments are expensive, Bayesian approaches are strongly preferred.

\section{Conclusion}
 We offer MOLTE as a public-domain test environment to facilitate the process of more comprehensive comparisons, on a broader set of test problems and a broader set of policies, so that researchers can more easily draw insights into the behavior of different policies in the context of different problem classes. There has been a long history in the optimal learning literature of proving some sort of bound, supported at times by relatively thin empirical work by comparing a few policies on a small number of randomly generated problems. When choosing policies from a huge algorithms pool, we hope \pkg{MOLTE} can be a starting point for researchers, experimental scientists and students to more easily draw insights into the behavior of different policies in the context of different problem classes.  We demonstrate the ability of \pkg{MOLTE} through extensive experimental results.  We draw the conclusion that there is no universal best policy for all problem classes, and bounds, by themselves, do not provide reliable guidance to the policy that will work the best.  We envision \pkg{MOLTE} as a modest spur to induce other researchers to come forward to study interesting questions involved in optimal learning, for example, the issue of tuning in this paper. We hope \pkg{MOLTE} can help with the current issue of relative paucity of empirical testing of learning algorithms.

\newpage

\footnotesize{
\bibliography{refer}}
\end{document}